\pdfoutput=1

\documentclass[11pt]{article}

\usepackage{acl}

\usepackage{times, soul}
\usepackage{latexsym}

\usepackage[T1]{fontenc}

\usepackage[utf8]{inputenc}

\usepackage{microtype}

\usepackage{inconsolata}

\usepackage{graphicx}

\usepackage{multicol}
\usepackage{multirow}
\usepackage{enumitem}
\usepackage{hyperref}
\usepackage{booktabs}
\usepackage{amssymb}
\usepackage{amsmath,mathtools,xspace,color,soul}
\usepackage{subcaption}
\usepackage{appendix}
\usepackage{algorithm}
\usepackage{algpseudocode}
\usepackage{marvosym}
\usepackage[english]{babel}

\DeclareMathOperator*{\argmax}{argmax}

\title{\method: Mastering Multi-turn Conversational Recommendation with Strategic Planning and Monte Carlo Tree Search}

\author{Hanwen Du\textsuperscript{$\clubsuit$} \  Bo Peng\textsuperscript{$\clubsuit$} \  Xia Ning\textsuperscript{$\clubsuit$}\textsuperscript{$\spadesuit$}\textsuperscript{$\heartsuit$}\Letter \\
\textsuperscript{$\clubsuit$}Department of Computer Science
and Engineering, The Ohio State University, USA\\
\textsuperscript{$\spadesuit$}Department of Biomedical Informatics, The Ohio State University, USA\\
\textsuperscript{$\heartsuit$}Translational Data Analytics Institute, The Ohio State University, USA\\
\texttt{\{du.1128,peng.707,ning.104\}@osu.edu}}

\newcommand{\method}{\mbox{$\mathop{\mathtt{SAPIENT}}\limits$}\xspace}
\newcommand{\agent}{\mbox{$\mathop{\mathtt{S}\text{-agent}}\limits$}\xspace}
\newcommand{\planner}{\mbox{$\mathop{\mathtt{S}\text{-planner}}\limits$}\xspace}

\newcommand{\methodeff}{\mbox{$\mathop{\mathtt{SAPIENT\text{-e}}}\limits$}\xspace}
\newcommand{\AbsGreedy}{\mbox{$\mathop{\mathtt{Abs{\thinspace\thinspace}Greedy}}\limits$}\xspace}
\newcommand{\MaxEntropy}{\mbox{$\mathop{\mathtt{Max{\thinspace\thinspace}Entropy}}\limits$}\xspace}
\newcommand{\CRM}{\mbox{$\mathop{\mathtt{CRM}}\limits$}\xspace}
\newcommand{\EAR}{\mbox{$\mathop{\mathtt{EAR}}\limits$}\xspace}
\newcommand{\SCPR}{\mbox{$\mathop{\mathtt{SCPR}}\limits$}\xspace}
\newcommand{\UNICORN}{\mbox{$\mathop{\mathtt{UNICORN}}\limits$}\xspace}
\newcommand{\MCMIPL}{\mbox{$\mathop{\mathtt{MCMIPL}}\limits$}\xspace}
\newcommand{\HutCRS}{\mbox{$\mathop{\mathtt{HutCRS}}\limits$}\xspace}
\newcommand{\CORE}{\mbox{$\mathop{\mathtt{CORE}}\limits$}\xspace}

\begin{document}
\maketitle
\begin{abstract}
Conversational Recommender Systems (CRS) proactively engage users in interactive dialogues to elicit user preferences and provide personalized recommendations. 
Existing methods train Reinforcement Learning (RL)-based agent with greedy action selection or sampling strategy, and may suffer from suboptimal conversational planning. To address this, we present a novel Monte Carlo Tree Search (MCTS)-based CRS framework \method.
\method consists of a conversational agent (\agent) and a conversational planner (\planner).
\planner builds a conversational search tree with MCTS based on the initial actions proposed by \agent to find conversation plans.
The best conversation plans from \planner are used to guide the training of \agent, creating a self-training loop where \agent can iteratively improve its capability for conversational planning.
Furthermore, we propose an efficient variant \methodeff for trade-off between training efficiency and performance. Extensive experiments on four benchmark datasets validate the effectiveness of our approach, showing that \method outperforms the state-of-the-art baselines. Our code and data are accessible through \url{https://github.com/ninglab/SAPIENT}.
\end{abstract}

\begin{figure*}[htbp]
    \centering
\includegraphics[width=\textwidth]{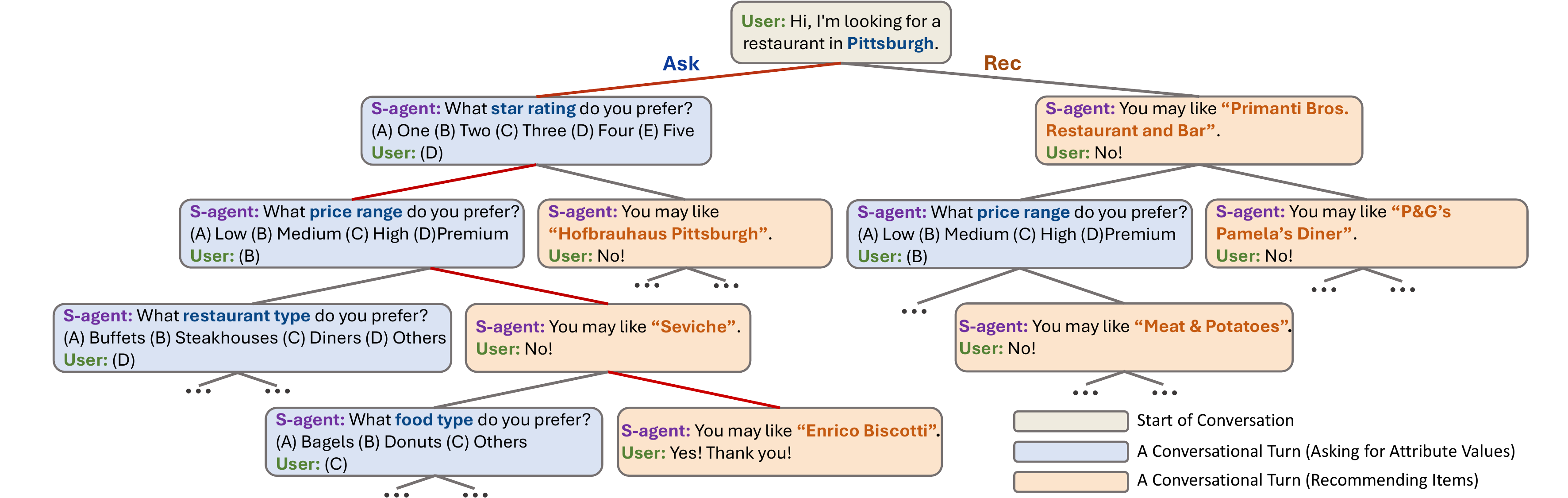}
    \caption{An example of conversational search tree for a user. Conversation starts at the root node with the user specifying preference on an attribute type and its value. The search tree expands as \agent decides different action types---$\mathtt{ask}$ and $\mathtt{rec}$---at each turn. Red line connects the highest-rewarded conversation plan found by the tree.}
    \label{CRSTree}
\end{figure*}
\section{Introduction}
Conversational Recommender Systems (CRSs) are developed to proactively engage users with interactive dialogues to understand user preferences and provide highly personalized recommendations \cite{christakopoulou2016towards,lei2020estimation}. 
For example, on an online dining platform such as Yelp \cite{lei2020estimation}, 
CRS can chat with users through natural language dialogues (e.g., ask a question like, ``What is your preferred food type?'') and recommend products that best match the users' preferences expressed in the conversation.
Among different settings of CRS \cite{sun2018conversational,deng2021unified,he2023large}, the Multi-turn Conversational Recommendation (MCR) setting~\cite{lei2020estimation,lei2020interactive,deng2021unified} is 
popular as it can interact/communicate with users multiple times (i.e., multiple turns) to iteratively learn
user preferences \cite{fu2020tutorial,jannach2021survey}.
In this work, we develop an innovative Monte Carlo Tree Search (MCTS)-based MCR framework to enhance the strategic conversational planning ability for MCR, offering a fresh perspective for handling complex conversation environments and enhancing user experiences.

A key of MCR is to decide what action (asking a question on specific attribute values or recommending specific items) to take at each conversational turn---a conversational turn consists of the CRS taking an action and the user responding to that action---to effectively elicit information on user preferences and make personalized recommendations \cite{fu2020tutorial,lei2020conversational}.
To achieve this, previous methods formulated MCR as a Markovian Decision Process (MDP) \cite{bellman1957markovian}, 
and trained policy-based \cite{sun2018conversational,lei2020estimation} or value-based \cite{deng2021unified,zhang2022multiple} agents via Reinforcement Learning (RL) to learn conversation strategies. 
Despite promising, 
these methods could suffer from myopic actions and limited planning capability due to the following reasons.
First, they base their planning solely on observations of the current state (e.g., items that the user indicates a negative preference for) without exploring potential future states. 
As a result, they could take myopic actions~\cite{anthony2017thinking,cohen2022dynamic}.
Second, they generate conversation trajectories, also referred to as conversation plans, by sequentially sampling actions, and thus could suffer from the cumulative error, especially when generating long trajectories for planning~\cite{aviral2019stabilizing,Lan2020Maxmin}.

To address these limitations, we present a novel MCTS-based MCR framework---\underline{S}trategic \underline{A}ction \underline{P}lanning with \underline{I}ntelligent \underline{E}xploration \underline{N}on-myopic \underline{T}actics, referred to as \method.
\method comprises a conversational agent, referred to as {\agent}, 
where {\agent} utilizes an MCTS-based algorithm, 
referred to as {\planner}, to plan conversations.
\agent builds a global information graph and two personalized graphs with dedicated graph encoders to extract the representation of the conversational states, and synergizes a policy network and a Q-network to decide specific actions based on the learned state representations.
\planner leverages MCTS \cite{kocsis2006bandit,coulom2007efficient} to simulate future conversations with lookahead explorations. This non-myopic conversational planning process ensures \planner can strategically plan conversations that maximize the cumulative reward (a numerical signal measuring whether the action taken by \agent is good or not), instead of greedily selecting actions based on immediate reward.
The best conversation plans with the highest cumulative rewards found by \planner are used to guide the training of \agent. 
In this way, \agent can engage in a self-training loop~\cite{silver2017mastering}---collecting trajectories from multiple conversation simulations and training on selected, high-rewarded trajectories---to iteratively improve its planning capability without additional labeled data.
After \agent is well-trained, it can directly make well-informed decisions without \planner during inference, since it inherits the \planner's expertise in strategic, non-myopic planning.

To make MCTS scalable w.r.t. the size of items and attributes, we introduce a hierarchical action selection process \cite{machum2018data}, and two action types: $\mathtt{ask}$ and $\mathtt{rec}$. 
At each turn, instead of searching over all the items and attribute values, {\planner} builds a conversational search tree (Figure~\ref{CRSTree}) that only searches over the two action types and uses the Q-network to decide the specific action, thus greatly reducing the search space.

We evaluate \method against 9 state-of-the-art CRS baselines, and show \method significantly outperforms baselines on 4 benchmark datasets.
Our case study also shows that the action strategies of \method are beneficial for information seeking and recommendation success in the conversations.

Furthermore, we develop an efficient variant of \method, denoted as \methodeff.
Different from \method, which is trained on selected, high-rewarded trajectories, \methodeff consumes all trajectories found by \planner for training via a listwise ranking loss.
As a result, \methodeff requires less cost of collecting training trajectories compared to \method, and enables superior efficiency.
Our contributions are summarized as follows:

\begin{itemize}[noitemsep,nolistsep,leftmargin=*]
   \item We present \method, a novel MCR framework synergizing an MCTS-based \planner and an \agent with a self-training loop to iteratively improve \agent's planning capability. To the best of our knowledge, \method is the first to leverage an MCTS-based planning algorithm to achieve strategic, non-myopic planning for MCR.
   \item We further develop \methodeff, an efficient variant trained on all trajectories from \planner via a listwise ranking loss. \methodeff
 addresses the efficiency issue with MCTS while maintaining similar performance with \method.
   \item Our extensive experiments show both \method and \methodeff outperform the state-of-the-art baselines. Our case study shows \method can strategically take actions that enhance information seeking and recommendation success.
\end{itemize}
\section{Related Work}
\paragraph*{Conversational Recommender System} CRS understands user preferences through interactive natural language conversations to provide personalized recommendations \cite{fu2020tutorial,jannach2021survey}. 
Early methods \cite{christakopoulou2016towards,sun2018conversational} ask users about their desired attribute values to narrow down the list of candidate items to recommend, but are limited under the single-turn setting, as they can only recommend once in a conversation. 
To address this, multi-turn CRSs allow for multiple turns of question inquiries and item recommendations. 
For example, \EAR \cite{lei2020estimation} adjusts the conversation strategy based on the user's feedbacks with a three-staged process. 
\SCPR \cite{lei2020interactive} models MCR as a path reasoning problem over the knowledge graph of users, items, and attribute values. 
\UNICORN \cite{deng2021unified} introduces a graph-based RL framework for MCR. 
\MCMIPL \cite{zhang2022multiple} develops a multi-interest policy learning framework to understand user's interests over multiple attribute values.
\HutCRS \cite{qian2023hutcrs} introduces a user interest tracing module to track user preferences. 
\CORE \cite{jin2023lending} designs a CRS framework powered by large language models with user-friendly prompts and interactive feedback mechanisms.
\citet{chen2019large} and \citet{Montazeralghaem2021large} build a tree-structured index with clustering algorithms to handle the large scale of items and attribute values in MCR.
\emph{Different from these methods, \method can iteratively improve its planning ability through self-training on demonstrations from MCTS, allowing for more informed and non-myopic conversation strategies.}
\paragraph*{Reinforcement Learning for CRS} Reinforcement Learning (RL) has achieved great success in tasks requiring strategic planning in complex and interactive environments, such as computer Go \cite{silver2016mastering,silver2017mastering} and dialogue planning \cite{yu-etal-2023-prompt,he2024planning}.
RL is also employed to train CRS agents to make strategic actions, and current RL-based CRSs can be mainly categorized into two types of methods: \textbf{(1)} policy-based methods, which train a policy network that directly outputs the probability of taking each action \cite{sun2018conversational,lei2020conversational}, and \textbf{(2)} value-based methods, which train a Q-network \cite{van2016deep} to estimate the Q-value of actions \cite{deng2021unified,zhang2022multiple}.
Despite promising, these CRS methods may suffer from myopic conversational planning and suboptimal decisions due to their greedy action selection and sampling strategy.
\emph{In contrast to these methods, our new \method is able to achieve strategic and non-myopic conversational planning through an MCTS-based planning and self-training algorithm.}
\section{Notations and Definitions}
We denote $\mathcal{U}$ as the set of users, $\mathcal{V}$ as the set of items, 
$\mathcal{Y}$ as the set of attribute types (e.g., price range, star rating), and 
$\mathcal{P}$ as the set of attribute values (e.g., medium price range, five-star rating).
Each user $u\in\mathcal{U}$ has an interaction history (e.g., view, purchase) with a set of items $\mathcal{V}(u)$.
Each item $v\in\mathcal{V}$ is associated with a set of attribute types $\mathcal{Y}(v)$ and the corresponding set of attribute values $\mathcal{P}(v)$.
%
Each conversation is initialized by a user specifying preference on an attribute type $y_{0}\in\mathcal{Y}$ and its corresponding attribute value $p_{0}\in\mathcal{P}$ (e.g., the user says ``I am looking for a place with medium price range.'').
At the $t$-th conversational turn, \agent can either ask for preferences over attribute values from a set of candidate attribute values $\mathcal{P}^{c}_{t}$, or recommend items from a set of candidate items $\mathcal{V}^{c}_{t}$.
Based on the user's reply (accept or reject attribute values/items), \agent repeatedly communicates with users until the user accepts at least one recommended item at turn $T$ (success), or the conversation reaches the maximum number of turns and terminates at $T{=}T_{\text{max}}$ (fail). 
The goal of MCR is to recommend at least one item that the user accepts, and complete the conversation in as few turns as possible to prevent the user from becoming impatient after too many turns.
\section{Method}
\begin{figure*}[t]
    \centering
    \includegraphics[width=\linewidth]{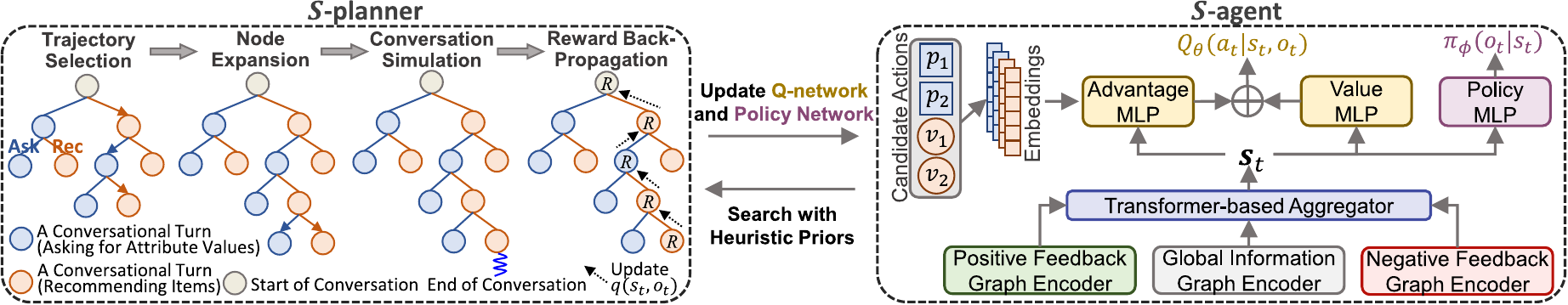}
    \caption{\method consists of a conversational agent (\agent) and a conversational planner (\planner). \planner leverages MCTS to perform non-myopic conversational planning based on the heuristics from \agent. The best conversation plans found by \planner are used to guide the training of \agent, enabling \agent to engage in a self-training loop that iteratively improves its capability for conversational planning.}
    \label{fig:overview}
\end{figure*}

We introduce \method, an MCTS-based MCR framework that achieves strategic and non-myopic conversational
planning.
\method formulates MCR as an MDP with a hierarchical action selection process (Section~\ref{sec:MDPFormulation}).
A conversational agent (\agent) observes the current state and decides the actions in each conversational turn (Section~\ref{sec:agent}), a conversational planner (\planner) leverages an MCTS-based algorithm to plan conversations (Section~\ref{sec:planner}), and \agent engages in a self-training loop with guidance from \planner (Section~\ref{sec:learningfrommctstrajectories}). Once \agent is well-trained, it can directly make well-informed decisions without \planner during inference, since it inherits the \planner's expertise in strategic and non-myopic planning. A framework overview of \method is in Figure~{\ref{fig:overview}}
and the training algorithm is in Algorithm~\ref{alg_overview:sapient}. We summarize all the notations in Appendix~\ref{sec:notation}.
\subsection{MDP Formulation for MCR}
\label{sec:MDPFormulation}

We formulate MCR as an MDP where \agent can be trained in an RL environment to learn to plan conversations strategically. 
For each user $u$, the MDP environment $\mathcal{M}(u)$ is defined as a quintuple 
$\mathcal{M}(u)=\{\mathcal{S}, \mathcal{A}, \mathcal{T}, \mathcal{R}, \gamma\}_u$ 
(index $u$ is dropped when no ambiguity arises), where $\mathcal{S}$ denotes the state space, which summarizes all the information about the conversation and the user; $\mathcal{A}$ denotes the action space, which includes asking ($\mathtt{ask}$) for specific attribute types and their respective values, or recommending ($\mathtt{rec}$) specific items; 
$\mathcal{T}:\mathcal{S}\times\mathcal{A}\rightarrow\mathcal{S}$ denotes the transition to the next state after taking an action from the current state; $\mathcal{R}:\mathcal{S}\times\mathcal{A}\rightarrow\mathbb{R}$ denotes the immediate reward function after taking an action at the current state; and $\gamma\in (0, 1)$ denotes the discount factor.
For hierarchical action selection~\cite{machum2018data}, 
\agent first chooses an action type $o_{t}\in\{\mathtt{ask},\mathtt{rec}\}$ at each conversational turn, indexed by $t$, 
then chooses the objective of that action type. In summary, the MDP environment provides information about the current state for \agent, and trains it to maximize the reward by optimizing its action strategy.
%
%
\paragraph*{State}
For the $t$-th turn, we define the state $s_{t}\in\mathcal{S}$ as a triplet $s_{t}=(\mathcal{P}^{+}_{t},\mathcal{P}^{-}_{t},\mathcal{V}^{-}_{t})$, where $\mathcal{P}^{+}_{t}$ denotes all the attribute values that the user has accepted until the $t$-th turn, $\mathcal{P}^{-}_{t}$ and $\mathcal{V}^{-}_{t}$ denote all the attribute values and items that the user has rejected until the $t$-th turn. 
As the conversation continues until the user accepts a recommended item, or the conversation terminates at $T_{\text{max}}$ turns when none of the recommended items are accepted by the user, the accepted item set $\mathcal{V}^{+}_{t}$ is always empty during the conversation, and hence we do not need $\mathcal{V}^{+}_{t}$ in the state. 
Besides, \agent also has access to all the information about the user $u$
and the global information graph $\mathcal{G}$ (a tripartite graph that represents all the interactions between users and items and all the associations between items and attribute
values). 
The state is initialized as $s_{0}$ when the user specifies preference on an attribute type $y_{0}\in\mathcal{Y}$ and its corresponding attribute value $p_{0}\in\mathcal{P}$,
and transitions to the next states as the conversation continues. 
The candidate attribute value set $\mathcal{P}^{c}_{t}$ and candidate item set $\mathcal{V}^{c}_{t}$ are updated according to $\mathcal{P}^{+}_{t}$, $\mathcal{P}^{-}_{t}$ and $\mathcal{V}^{-}_{t}$, which we will elaborate later 
in the \textbf{Transition} subparagraph. We present an illustration on how to calculate the state $s_t$ in Appendix~\ref{sec:state_illustration}.
\paragraph*{Action}
The action $a_t$ refers to asking for a specific attribute value ($\mathtt{ask}$) 
or recommending a specific item ($\mathtt{rec}$) at the $t$-th turn. 
Here, we adopt a hierarchical action selection process: 
we first use a new policy network $\pi_{\phi}(o_{t}|s_t)$ to decide the action type $o_{t}\in\{\mathtt{ask},\mathtt{rec}\}$ from the current state $s_t$, 
and then use a new Q-network $Q_{\theta}(a_t| s_t, o_t)$ to decide the specific action $a_t$ according to the action type $o_t$.
The action space (at the current state $s_t$) $\mathcal{A}_{s_t}{=}\{\mathcal{P}^{c}_{t},\mathcal{V}^{c}_{t}\}$ contains all the candidate items and attribute values.
%
The Q-network $Q_{\theta}(a_t| s_t, o_t)$  only selects an action from a sub action space $\mathcal{A}_{s_t, o_{t}}$: when $o_{t}{=}\mathtt{ask}$, $\mathcal{A}_{s_t, o_{t}}{=}\mathcal{P}^{c}_{t}$; when $o_{t}{=}\mathtt{rec}$, $\mathcal{A}_{s_t, o_{t}}{=}\mathcal{V}^{c}_{t}$. Details on the policy network and the Q-network are available in Section~\ref{sec:agent}.
\paragraph*{Transition}
Transition occurs from the current state $s_t$ to the next state $s_{t+1}$ when the user responds 
to the action $a_t$ (accepts or rejects items/attribute values). 
Candidate item set are narrowed down to the remaining items that still satisfy the user's preference requirement, and attribute values asked at turn $t$ are excluded from the candidate attribute value set. More details are in Appendix~\ref{transitiondetails}.
\paragraph*{Reward}
We denote the immediate reward at the $t$-th conversational turn as $r_{t}$, and the cumulative reward for each conversation is calculated as $\sum^{T}_{t=1}\gamma^{t}r_t$.
Intuitively, a positive reward is assigned when the user accepts the items or attribute values, and a negative reward is assigned when the user rejects the items or attribute values. Details on the reward function are available in Appendix~\ref{detailreward}.
\subsection[S-agent]{\agent}
\label{sec:agent}
%
\agent comprises three components: the state encoder, the policy network, and the Q-network. The state encoder adopts graph neural networks to generate the state representation. This state representation is then utilized by both the policy network and the Q-network to decide the action type and the specific actions in each conversational turn.
\paragraph*{State Encoder} 
To include essential information about the conversation and the user, \agent first uses a graph attention network~\cite{brody2022how} to learn the representations of users, items, and attribute values out of the global information graph $\mathcal{G}$. 
Next, \agent introduces two personalized graphs---positive feedback graph (denoted as $\mathcal{G}^{+}_{t}$) and negative feedback graph (denoted as $\mathcal{G}^{-}_{t}$)---to represent each user's acceptance/rejection on attribute values until the $t$-th turn, and uses two dedicated graph convolutional networks to learn the representations of users, items, and attribute values~\cite{zhang2022multiple} from the graphs.
Finally, \agent aggregates the representations of items and attribute values from $\mathcal{G}$, $\mathcal{G}^{+}_{t}$ and $\mathcal{G}^{-}_{t}$ with a Transformer~\cite{vaswani2017attention}-based aggregator to model the action sequence in the conversation, and obtain the representation $\mathbf{s}_{t}$ of the current state $s_t$. Details on the state encoder are available in Appendix~\ref{technicaldetails_state_encoder}.
\paragraph*{Policy Network \& Q-Network}
Based on the state representation $\mathbf{s}_{t}$, \agent adopts a policy network $\pi_{\phi}(o_{t}|s_t)$ to decide action type $o_t$ and a Q-network $Q_{\theta}(a_t|s_t,o_t)$ to decide the specific action $a_t$ according to the action type $o_t$:
\begin{equation}
    \begin{gathered}
    \pi_{\phi}(o_{t}|s_t)=\text{softmax}(\text{MLP}_{\pi}(\mathbf{s}_{t}))\\
    Q_{\theta}(a_t| s_t, o_t)=\text{MLP}_{A}(\mathbf{s}_{t}|| \mathbf{a}_{t})+\text{MLP}_{V}(\mathbf{s}_{t}),
    \end{gathered}
\label{policyandqnetworkequation}
\end{equation}
where MLP denotes a two-layer perceptron, $\mathbf{a}_{t}=\mathbf{e}_{p}$ or $\mathbf{a}_{t}=\mathbf{e}_{v}$ denotes the embedding of actions (attribute value or item) at the $t$-th turn, $A$ and $V$ denote the advantage and value function of the dueling Q-network \cite{wang2016dueling} respectively.
\subsection[S-planner]{\planner}
\label{sec:planner}
%
\planner adopts an MCTS-based planning algorithm to simulate conversations and finds the best conversation plan for each user, strategically balancing exploration and exploitation to efficiently expand a search tree~\cite{kocsis2006bandit,coulom2007efficient}. Specifically, each node in the tree represents a state $s_t$, the root node $s_{0}$ represents the initial state where the user specifies preference on an attribute type and its corresponding value, and the leaf node represents the end of the conversation (success or fail). Each edge between nodes $s_t$ and $s_{t+1}$ represents an action type $o_t\in\{\mathtt{ask}$, $\mathtt{rec}\}$ and the transition from the current state $s_t$ to the next state $s_{t+1}$ after choosing an action type $o_t$ and a specific action $a_t$. For each action type $o_t$, \planner maintains a function $q(s_t, o_t)$ of $s_t$ and $o_t$ as the expected future reward of selecting action type $o_t$ at the state $s_t$. For each user $u$, \planner simulates $N$ different conversation plans (also referred to as trajectories), and the trajectory for the $i$-th simulation is denoted as $\tau_{i}^{(u)}$, which contains a sequence of state $s_t$, action type $o_t$, action $a_t$ and immediate reward $r_t$ at each conversational turn $t$. 
The search tree is built in four stages: 
\begin{itemize}[noitemsep,nolistsep,leftmargin=*]
    \item Trajectory selection: 
	\planner traverses from the root to leaves over the current tree to select the most promising trajectory that is likely to obtain a high cumulative reward.
    \item Node expansion:
    	\planner initializes two children nodes ($\mathtt{ask}$ and $\mathtt{rec}$) to the leaf node 
	on the selected trajectory to expand the tree.
    \item Conversation simulation: 
    \planner simulates future conversations between \agent and the user, starting from the expanded node and foresees how the future conversation will unfold.
    \item Reward back-propagation: \planner updates the expected future reward $q(s_t, o_t)$ of action type $o_t$ along the trajectory using the cumulative reward of the current conversation.
\end{itemize}
%
\paragraph*{Trajectory Selection}
\planner selects the most promising trajectory from the root node to a leaf node 
that is likely to obtain a high future reward $q(s_t, o_t)$, and the selected trajectory will be further expanded for conversation simulation later. 
This selection process trades off between exploitation, measured by $q(s_t, o_t)$, 
against exploration, measured by how often the nodes are visited. 
Particularly, \planner adapts the Upper Confidence bounds applied to Trees (UCT) approach~\cite{kocsis2006bandit} to achieve the trade-off between exploitation and exploration, 
and at each node $s_t$, select action type $o^*_t$ into the trajectory that maximizes the UCT value as follows:
%
\begin{equation}
\label{eqn:UCT}
o^*_t \gets\argmax\limits_{o_{t}\in\{\mathtt{ask},\mathtt{rec}\}}
\bigg[q(s_t,o_{t})+w\sqrt{\frac{\log V(s_t)}{V(f(s_t, o_t))}}\bigg],
\end{equation}
where $w>0$ is the exploration factor,
$V(s_t)$ quantifies the visits on node $s_t$ during conversation simulations, and 
$f(s_t, o_t)$ represents the child node of $s_t$ after choosing the action type $o_t$.
Intuitively, the second term in Equation~\ref{eqn:UCT} is larger if the child node is less visited, 
encouraging more exploration.
After selecting the action type $o_t^*$, \planner chooses the optimal action $a^*_{t}$ with the Q-network as follows:
\begin{equation}
\label{eqn:UCT_act}
a^*_t \gets \argmax_{a_{t}\in\mathcal{A}_{s_{t}, o^*_{t}}}Q_{\theta}(a_t|s_t, o^*_t).
\end{equation}
%
\paragraph*{Node Expansion}
%
When a leaf node is reached, 
\planner expands the leaf node by attaching two children nodes (corresponding to two action types $\mathtt{ask}$ and $\mathtt{rec}$) to it. 
The expected future reward $q(s_{t+1}, o_{t+1})$ of choosing $o_{t+1}$ at the newly attached node $s_{t+1}$ is initialized as the highest value estimated by the Q-network $Q_{\theta}(a_{t+1}|s_{t+1}, o_{t+1})$
among all the candidate actions in the sub action space $\mathcal{A}_{s_{t+1}, o_{t+1}}$, serving as a heuristic guidance for the future tree search.

\paragraph*{Conversation Simulation}
%
To predict how the future conversation unfolds, \planner continues to simulate conversations between \agent and the user until the conversation succeeds or fails. Starting from the last expanded node, at each turn, the policy network decides the action type, while the Q-network decides the specific action.
\paragraph*{Reward Back-Propagation}
%
Once the simulated conversation succeeds or fails, \planner back-propagates from the leaf node of the current trajectory $\tau^{(u)}_{i}$ to the root to increase the visit count of each node along $\tau^{(u)}_{i}$, and update the expected future reward $q(s_t,o_{t})$ along $\tau^{(u)}_{i}$ as follows:
\begin{equation}
\label{eqn:backprop}
q(s_t,o_{t}){\gets}q(s_t,o_{t})+\big({R_{t}(\tau^{(u)}_{i})-q(s_t,o_{t})}\big)/{V(s_t)},
\end{equation}
where $R_{t}(\tau^{(u)}_{i})=\sum^{T}_{\hat{t}=t}\gamma^{\hat{t}-t}{r}_{\hat{t}}$ is the cumulative reward from turn $t$ to the final turn $T$.
Intuitively, this update rule is similar to stochastic gradient ascent: at each iteration the value of $q(s_t,o_{t})$ is adjusted by step $1/{V(s_t)}$ in the direction of the error $R_{t}(\tau^{(u)}_{i})-q(s_t,o_{t})$.

\subsection[Guiding S-agent with S-planner]{Guiding \agent with \planner}
\label{sec:learningfrommctstrajectories}
To empower \agent with advanced planning capability, we use the best conversation plan (the plan with the maximum cumulative reward) found by \planner to guide the training of its policy network and the Q-network.
This process creates a self-training loop~\cite{silver2017mastering} that enables \agent to iteratively improve its planning capability.
To avoid biased estimation from training on consecutive, temporally correlated actions~\cite{mnih2015human}, we store the experiences $e_t=(s_t, o_t,a_t,r_t,s_{t+1}, o_{t+1})$ at each turn $t$ ($s_{t+1}$ and $o_{t+1}$ are required for the target Q-network to estimate the Q value from the next state) from the best plans 
to the memory $\mathcal{D}$, and use Prioritized Experience Replay (PER) \cite{SchaulQAS15} to sample a batch of experiences from the memory $\mathcal{D}$ to update the policy network and the Q-network.
\begin{algorithm}[t]
\small
\caption{Training algorithm of \method}
\label{alg_overview:sapient}
\begin{algorithmic}
\Require conversational MDPs for all users $\{\mathcal{M}(u)\}^{\mathcal{U}}_{u=1}$, training steps $E$, \# of simulations $N$, exploration factor $w$
\For {$\text{step}\gets 1,\cdots, E$}
\State Sample a user $u$ from $\mathcal{U}$, initialize the state as $s_0$
    \For{$n\gets 1, \cdots, N$}
        \State Initialize the trajectory as $\tau^{(u)}_{i}\gets\{\}$, $t\gets0$
        \While{$s_t$ has children}\Comment{Trajectory Selection}
            \State Select an action type $o_{t}$ (Eq.~\ref{eqn:UCT})
            \State Select an action $a_t$ (Eq.~\ref{eqn:UCT_act})
            \State Save $s_t, o_t, a_t, r_t$ to $\tau^{(u)}_{i}$ \State $s_{t+1}\gets \mathcal{T}(s_t,a_t), t\gets t+1$
        \EndWhile
        \While{$s_t$ is not end of conversation}
        
        \Comment{Node Expansion}
        \State Attach two children ($\mathtt{ask}$ and $\mathtt{rec}$) to $s_t$ 
            %
            \State\Comment{Conversation Simulation}
            \State Select $o_t$ with $\pi_{\phi}(o_{t}|s_t)$, $a_t$ with $Q_{\theta}(a_t|s_t, o_t)$
            \State Save $s_t, o_t, a_t, r_t$ to $\tau^{(u)}_{i}$ \State $s_{t+1}\gets\mathcal{T}(s_t, a_t)$, $t\gets t+1$
        \EndWhile
        \State Initialize $R_t(\tau^{(u)}_{i})\gets 0$
        \While{$t \geq 0$}\Comment{Reward Back-Propagation}
            \State $R_{t}(\tau^{(u)}_{i}){\gets}{\gamma}R_{t}(\tau^{(u)}_{i})+r_t$, $V(s_t){\gets} V(s_t)+1$
            \State Update $q(s_t, o_t)$ with Eq.~\ref{eqn:backprop}, $t\gets t-1$
        \EndWhile
    \EndFor\Comment{Training}
    \State Save the highest-rewarded trajectory to the memory $\mathcal{D}$
    \State Sample $e_t\sim\mathcal{D}$, update $\pi_{\phi}(o_{t}|s_t)$, $Q_{\theta}(a_t| s_t, o_t)$
\EndFor
%
\end{algorithmic}
\end{algorithm}
\paragraph*{Policy Network Update}
The policy network is updated with the following supervised loss function to align its decision with guidance from \planner:
\begin{equation}
\label{eqn:policy}
\mathcal{L}_{\phi}=\mathbb{E}_{e_t\sim\mathcal{D}}\left[-\log{\pi_{\phi}(o_{t}|s_t)}\right].
\end{equation}
\paragraph*{Q-Network Update}
The Q-network is updated with double Q-learning \cite{van2016deep}, which maintains a target network ${Q}_{\tilde{\theta}}(a_t| s_t, o_t)$ as a periodic copy of the online network $Q_{\theta}(a_t| s_t, o_t)$ and trains $Q_{\theta}(a_t| s_t, o_t)$ to minimize the temporal difference error \cite{sutton1988learning}:
\begin{equation}
\label{eqn:qnet}
\begin{aligned}
&\mathcal{L}_{\theta}=\mathbb{E}_{e_t\sim\mathcal{D}}\Big[\big(Q_{\theta}(a_t| s_t, o_t)-r_t\\&-\gamma\max_{a_{t+1}\in\mathcal{A}_{s_{t+1}, o_{t+1}}}Q_{\tilde{\theta}}(a_{t+1}| s_{t+1}, o_{t+1})\big)^{2}\Big],
\end{aligned}
\end{equation}
where $\gamma$ is the discount factor in MDP. 

\paragraph*{Improving Training Efficiency}
%
The aforementioned training process guarantees the quality of training data by selecting only the high-rewarded trajectories, but may be inefficient because it requires a large number of simulations to collect enough high-rewarded trajectories. 
To improve efficiency, we further propose a variant \methodeff. Instead of using only the highest-rewarded trajectories, \methodeff makes full use of all the trajectories found by \planner. As \agent usually requires fixed number of training trajectories to converge, utilizing all the trajectories---rather than just a selected few---greatly reduces the cost of collecting trajectories and improves training efficiency.

Since some trajectories are good while others are suboptimal (e.g., user quits the conversation after $T_{\text{max}}$ turns), we would like to encourage $\pi_{\phi}(o_t|s_t)$ to increase the likelihood for good trajectories and decrease the likelihood for the suboptimal ones. 
To this end, we employ the Plackett-Luce model \cite{luce1959individual,plackett1975analysis} to train $\pi_{\phi}(o_t|s_t)$ with listwise likelihood estimations. For each user $u$, assuming all the $N$ trajectories are ranked by their cumulative rewards in the order of $\tau^{(u)}_1, \tau^{(u)}_2, \cdots, \tau^{(u)}_N$, the policy network is updated with the following loss function:
\begin{align}
\mathcal{L}_{\phi}=\mathbb{E}_{u\sim\mathcal{U}}\left[-\log P(\tau_1^{(u)}\succ\tau_2^{(u)}\succ\cdots\succ\tau_N^{(u)})\right]\nonumber\\
=\mathbb{E}_{u\sim\mathcal{U}}\Bigg[-\log\prod\limits^{N}_{n=1}\frac{\exp\big(\sum\limits_{\mathclap{\substack{s_t,o_t\in\tau_{n}^{(u)}}}}\log\pi_{\phi}(o_t|s_t)\big)}{\sum\limits^{N}_{j=n}\exp\big(\sum\limits_{\mathclap{\substack{s_t,o_t\in\tau_{j}^{(u)}}}}\log\pi_{\phi}(o_t|s_t)\big)}\Bigg],
\end{align}
where $\tau_1^{(u)}\succ\tau_2^{(u)}$ indicates $\tau_1^{(u)}$ has higher cumulative reward than $\tau_2^{(u)}$,
and the denominator sums the likelihood of all the trajectories with higher cumulative reward than the $j$-th trajectory. 
The Q-network is still updated as in Equation~\ref{eqn:qnet} except that the sampled experiences come from all the trajectories instead of only the highest-rewarded trajectories. In this way, all the trajectories found by \planner can be utilized, thus saving the search cost. 
\methodeff only performs slightly worse than \method and much better than baselines (Section~\ref{sec:overall}), and can be viewed as a good trade-off between efficiency and performance.
\begin{table*}[htbp]
    \centering
    \small
\setlength{\tabcolsep}{3pt}{
    \begin{tabular}{llllllllllllllll}
    \toprule
    \multirow{2.5}{*}{\textbf{Models} }&\multicolumn{3}{c}{\textbf{Yelp}}&&\multicolumn{3}{c}{\textbf{LastFM} }&&\multicolumn{3}{c}{\textbf{Amazon-Book} }&&\multicolumn{3}{c}{\textbf{MovieLens}}\\
    \cmidrule{2-4}
   \cmidrule{6-8}
    \cmidrule{10-12}
    \cmidrule{14-16}
    &\multicolumn{1}{c}{SR$\uparrow$}&\multicolumn{1}{c}{AT$\downarrow$}&\multicolumn{1}{l}{\hspace{-0.1cm}hDCG$\uparrow$}&&\multicolumn{1}{c}{SR$\uparrow$}&\multicolumn{1}{c}{AT$\downarrow$}&\multicolumn{1}{l}{\hspace{-0.1cm}hDCG$\uparrow$}&&\multicolumn{1}{c}{SR$\uparrow$}&\multicolumn{1}{c}{AT$\downarrow$}&\multicolumn{1}{l}{\hspace{-0.1cm}hDCG$\uparrow$}&&\multicolumn{1}{c}{SR$\uparrow$}&\multicolumn{1}{c}{AT$\downarrow$}&\multicolumn{1}{l}{\hspace{-0.1cm}hDCG$\uparrow$}\\
    \midrule
    
    \AbsGreedy  &0.195&14.08&0.069&&0.539&10.92&0.251&&0.214&13.50&0.092&&0.752& 4.94&0.481\\
    
    \MaxEntropy    &0.375&12.57&0.139&&0.640&\phantom{0}9.62&0.288&&0.343&12.21&0.125&&0.704& 6.93&0.448\\
    
    \CRM         &0.223&13.83&0.073&&0.597&10.60&0.269&&0.309&12.47&0.117&&0.654&7.86&0.413\\
    
    \EAR         &0.263&13.79&0.098&&0.612&\phantom{0}9.66&0.276&&0.354&12.07&0.132&&0.714&6.53&0.457\\
    
    \SCPR        &0.413&12.45&0.149&&0.751&\phantom{0}8.52&0.339&&0.428&11.50&0.159&&0.812&4.03&0.547\\
    
    \UNICORN     &0.438&12.28&0.151&&0.843&\phantom{0}7.25&0.363&&0.466&11.24&0.170&&0.836&3.82&0.576\\
    
    \MCMIPL      &0.482&11.87&0.160&&0.874&\phantom{0}\underline{6.35}&\underline{0.396}&&0.545&10.83&0.223&&0.882&\underline{3.61}&\underline{0.599}\\

    \HutCRS      &\underline{0.528}&\underline{11.33}&\underline{0.175}&&\underline{0.900}&\phantom{0}6.52&0.348&&\underline{0.638}&\phantom{0}\underline{9.84}&\underline{0.227}&&\underline{0.902}&4.16&0.475\\

    \CORE      &0.210&12.82&0.166&&0.862&\phantom{0}7.05&0.356&&0.462&11.49&0.182&&0.810&6.51&0.429\\
    \midrule
    
    \methodeff     &0.612$^{*}$&10.41$^{*}$&0.208$^{*}$&&0.922$^{*}$&\phantom{0}6.32&0.358&&0.682$^{*}$&\phantom{0}9.51$^{*}$&0.239$^{*}$&&0.928$^{*}$&3.76&0.607$^{*}$\\
    
    \method     &\textbf{0.622}$^{*}$&\textbf{10.02}$^{*}$&\textbf{0.229}$^{*}$&&\textbf{0.928}$^{*}$&\phantom{0}\textbf{6.15}$^{*}$&\textbf{0.398}&&\textbf{0.718}$^{*}$&\phantom{0}\textbf{9.28}$^{*}$&\textbf{0.252}$^{*}$&&\textbf{0.930}$^{*}$&\textbf{3.48}$^{*}$&\textbf{0.610}$^{*}$\\
    \bottomrule
    \end{tabular}}
    \caption{Performances on four benchmark datasets. The best performance of our method and the best baseline in each column is in bold and underlined respectively. $*$ indicates that the improvement over the best baseline is statistically significant ($p<0.01$).}
    \label{results}
\end{table*}

\section{Experimental Settings}
%
\paragraph*{Datasets}
%
We evaluate \method on 4 benchmark datasets: Yelp \cite{lei2020estimation}, LastFM \cite{lei2020estimation}, Amazon-Book \cite{mcauley2015image} and MovieLens \cite{harper2015movielens}. 
Dataset details are available in Appendix~\ref{datasetstatistics}.

\paragraph*{User Simulator}
%
Training and evaluating CRS with real-world user interactions can be impractically expensive at scale. To address this issue, we adopt the user simulator approach~\cite{lei2020estimation} and simulate a conversation for each user as detailed in Appendix~\ref{detailusersimulator}. 
Note that this user simulator is widely adopted in the literature~\cite{lei2020estimation,deng2021unified,zhang2022multiple,zhao2023multi,qian2023hutcrs} and studies show the simulations are of high quality and suitable for evaluation purposes~\cite{lei2020conversational,zhang2020evaluating,zhang2022multiple}, allowing for large-scale evaluations at a relatively low cost.
\paragraph*{Evaluation Metrics}
Following the literature \cite{deng2021unified,zhang2022multiple}, the
Success Rate (SR) is adopted to measure the ratio of successful recommendations within $T_{\text{max}}$ turns; Average Turn (AT) to evaluate the average number of conversational turns; and hDCG \cite{deng2021unified} to evaluate the ranking order of the ground-truth item among the list of all the recommended items. 
For SR and hDCG, a higher value indicates better performance, while for AT, a lower value indicates better performance. 
Details for hDCG calculation are available in Appendix~\ref{hdcgdetail}.

\paragraph*{Baselines and Implementation Details}
%
We choose 9 state-of-the-art baselines for a comprehensive evaluation, including: (1) Max Entropy \cite{lei2020estimation}; (2) Abs Greedy \cite{christakopoulou2016towards}; (3) CRM \cite{sun2018conversational}; (4) EAR \cite{lei2020estimation}; (5) SCPR \cite{lei2020interactive}; (6) UNICORN \cite{deng2021unified}; (7) MCMIPL \cite{zhang2022multiple}; (8) HutCRS \cite{qian2023hutcrs};  and a Large Language Model (LLM) baseline CORE \cite{jin2023lending}. Baseline details are available in Appendix~\ref{baselinedetail}. Implementation details of \method are available in Appendix~\ref{implementationdetailourmethod}.
\section{Experimental Results}
\subsection{Overall Performance Comparison}
\label{sec:overall}
We compare \method with 9 state-of-the-art baselines and report the experimental results in Table~\ref{results}. We have the following observations: 

\textbf{(1)} \textbf{\textit{\method achieves consistent improvement over baselines in terms of all metrics on all the datasets, with an average improvement of 9.1\% (SR), 6.0\% (AT) and 11.1\% (hDCG) compared with the best baseline.}}
Different from baselines, which base their planning solely on the observation of current state without looking ahead, \method foresees how the future conversation unfolds with an MCTS-based planning algorithm. This enables \method to take actions that maximize the cumulative rewards instead of settling for the immediate reward, enabling strategic, non-myopic conversational planning and superior performances.

\textbf{(2)} \textbf{\textit{\method substantially outperforms baselines in datasets demanding strong strategic planning capability from the CRS.}} The performance gain of \method is higher on datasets with a larger AT (Yelp and Amazon-Book) compared to datasets with a smaller AT (LastFM and MovieLens), and higher AT in these datasets indicates the need for more strategic planning over long conversational turns. 
Compared with baselines, \method is equipped with \planner and excels in conversational planning, hence showing remarkable improvements on these two datasets.

\textbf{(3)} \textbf{\textit{\methodeff outperforms all baselines on recommendation success rate.}} Although the training data for \methodeff still contain a portion of low-quality trajectories, \methodeff still significantly outperforms the best baselines across most metrics, indicating that \methodeff is a good trade-off between efficiency and performance.
\subsection{Efficiency Analysis}
\begin{table*}[htbp]
\small
    \centering
      \setlength{\tabcolsep}{3pt}{
    \begin{tabular}{llllllllllllllll}
    \toprule
    \multirow{2.5}{*}{\textbf{Models} }&\multicolumn{3}{c}{\textbf{Yelp}}&&\multicolumn{3}{c}{\textbf{LastFM} }&&\multicolumn{3}{c}{\textbf{Amazon-Book} }&&\multicolumn{3}{c}{\textbf{MovieLens}}\\
    \cmidrule{2-4}
   \cmidrule{6-8}
    \cmidrule{10-12}
    \cmidrule{14-16}
    &\multicolumn{1}{c}{SR$\uparrow$}&\multicolumn{1}{c}{AT$\downarrow$}&\multicolumn{1}{l}{\hspace{-0.1cm}hDCG$\uparrow$}&&\multicolumn{1}{c}{SR$\uparrow$}&\multicolumn{1}{c}{AT$\downarrow$}&\multicolumn{1}{l}{\hspace{-0.1cm}hDCG$\uparrow$}&&\multicolumn{1}{c}{SR$\uparrow$}&\multicolumn{1}{c}{AT$\downarrow$}&\multicolumn{1}{l}{\hspace{-0.1cm}hDCG$\uparrow$}&&\multicolumn{1}{c}{SR$\uparrow$}&\multicolumn{1}{c}{AT$\downarrow$}&\multicolumn{1}{l}{\hspace{-0.1cm}hDCG$\uparrow$}\\
    \midrule    
    \method &\textbf{0.622}&\textbf{10.02}&\textbf{0.229}&&\textbf{0.928}&\textbf{6.15}&\textbf{0.398}&&\textbf{0.718}&\textbf{~~9.28}&\textbf{0.252}&&\textbf{0.930}&\textbf{3.48}&\textbf{0.610}\\\midrule
    w/o Global $\mathcal{G}$&0.520&11.39&0.171&&0.906&6.56&0.345&&0.626&10.15&0.217&&0.878&5.30&0.397\\
    w/o Positive $\mathcal{G}^{+}$&0.482&11.56&0.163&&0.862&7.53&0.313&&0.560&11.15&0.184&&0.886&4.17&0.496\\
    w/o Negative $\mathcal{G}^{-}$&0.532&10.80&0.185&&0.905&6.92&0.336&&0.656&10.01&0.227&&0.860&5.42&0.389\\\midrule
    w/o Pol. net.&0.519&11.08&0.186&&0.894&6.37&0.361&&0.628&~~9.61&0.240&&0.896&4.59&0.516\\
    w/o Q-net.&0.582&10.69&0.190&&0.808&7.92&0.332&&0.594&10.62&0.198&&0.866&5.47&0.386\\\midrule
    w/o \planner &0.520&11.06&0.193&&0.902&6.80&0.335&&0.650&10.20&0.218&&0.860&5.53&0.396\\
    \bottomrule
    \end{tabular}}
    \caption{Ablation studies on benchmark datasets. The best performance in each column is in bold.}
    \label{ablation_full}
\end{table*}
\begin{table}[ht]
    \centering
    \small
    \setlength{\tabcolsep}{1pt}{\begin{tabular}{lrrrr}
    \toprule    \textbf{Model}&\textbf{Yelp}&\textbf{LastFM}&\textbf{Amazon-Book}&\textbf{MovieLens}\\
    \midrule
    \UNICORN&16.15&4.30&6.03&7.96\\
    \MCMIPL&15.57&5.08&6.40&7.93\\
    \HutCRS&14.05&4.66&5.83&8.40\\
    \methodeff&16.40&5.57&6.88&8.45\\
    \method&38.15&11.07&13.21&20.97\\
    \bottomrule
    \end{tabular}}
    \caption{Training GPU hours on four datasets.}
    \label{traininghours}
\end{table}
\textit{\textbf{Training efficiency of \method and \methodeff is highly comparable to the baselines.}}
As shown in Table~\ref{traininghours} (all experiments are conducted on a single Tesla V100 GPU), \methodeff takes similar training time with baselines because it collects all the trajectories from MCTS and do not incur additional search cost. 
Even with \method, the training time is only about 2 times longer than baselines. This is because conversation simulation only requires forward propagation without gradient backward, so even conducting 20 rollouts per user will not significantly reduce efficiency.
Also note that during inference, the efficiency of \method is comparable with baselines, because tree search is not required during inference.
%
\subsection{Ablation Study}\label{sec:ablationstudy}
To validate the effectiveness of the key components in \method, we conduct ablation studies and report the results in Table~\ref{ablation_full}. From the experimental results, we have the following observations:

\textbf{(1)} \textit{\textbf{Each graph---$\mathcal{G},\mathcal{G}^{+},\mathcal{G}^{-}$ is vital for \agent to encode the state information.}} Removing each graph from \agent degrades performance, verifying the necessity of each graph in state encoding: global information graph $\mathcal{G}$ is crucial for mining user-item relations and item-attribute value associations, while positive ($\mathcal{G}^{+}$) and negative ($\mathcal{G}^{-}$) feedback graphs are vital for capturing users' preferences (likes/dislikes on items and attribute values) expressed in the conversation.

\textbf{(2)} \textit{\textbf{Both the policy network and the Q-network are critical to conversational planning.}} We design two variants: replacing the policy network with random action type selection (w/o Pol. net.); replacing the Q-network with entropy-based action selection (w/o Q-net.). Performance drops in both variants suggest both networks are crucial for hierarchical action selection, and the absence of an informed decision maker, either at the action type or the action level, leads to suboptimal conversational planning.

\textbf{(3)} \textit{\textbf{Guidance from \planner is crucial for \agent to achieve strategic conversational planning.}}
Removing \planner and training \agent on sampled on-policy trajectories as in~\citet{deng2021unified} degrades the performance, because sampled trajectories may bring cumulative errors and biased estimations~\cite{aviral2019stabilizing,Lan2020Maxmin}, resulting in suboptimal conversational planning. By contrast, the high-rewarded conversation plans from \planner offers robust guidance for \agent and boosts its capability for strategic planning.
%
\subsection{Hyperparameter Sensitivity}
\label{sec:hyperparameter_sensitivity}
We study \method's performance sensitivity to the exploration factor $w$ and the rollout number $N$, as detailed in Appendix~\ref{hyperparameter_sensitivity}. Our major conclusions are: the performance remains robust to large $w$ but drops with small $w$. \textit{\textbf{This suggests that \method favours exploration over exploitation during conversational tree search.}} Additionally, the performance notably improves when increasing from $N{=}1$ to $N{=}20$, and remains stable and satisfactory after $N{>}20$. \textit{\textbf{This suggests that setting $N{=}20$ can strike a good balance between efficiency (small $N$) and performance (large $N$).}}
%
\subsection{Case Study}
To gain an insight into the conversational planning capability of \method, we provide an analysis on the action strategies of \method (Appendix~\ref{actionstrategy}) and a case study (Appendix~\ref{casestudy}) to show \method can strategically take actions that are helpful for information seeking and recommendation success.
%
\section{Conclusion}
We present \method, a novel MCR framework with strategic and non-myopic conversational planning tactics. \method adopts a hierarchical action selection process, builds a conversational search tree with MCTS, and selects the high-rewarded conversation plans to train \agent. During inference, \agent can make well-informed decisions without \planner, as it inherits \planner's expertise in strategic planning. Furthermore, we develop a variant \methodeff to address the efficiency issue with MCTS. Extensive experiments on benchmark datasets verify the effectiveness of our framework.
\section{Limitations}
\paragraph*{Limited Action Types} Our framework only supports searching over two types of actions ($\mathtt{ask}$ and $\mathtt{rec}$) so far, which cannot search at a more fine-grained level (e.g., defining action types as ``recommending items with a five-star rating'', ``recommending items with a three-star rating'', rather than just ``recommending items''). For future work, we plan to adopt advanced action abstraction techniques \cite{bai2016marchovian} to divide the search space at more fine-grained levels. 

 \paragraph*{Training Cost} Conducting multiple simulated rollouts for each user with MCTS ensures the quality of the conversation plan, but also brings additional computational cost and reduces training efficiency. In the future, we plan to further improve the training efficiency of \method through techniques such as parallel acceleration \cite{chaslot2008parallel} for MCTS. 

\paragraph*{User Simulator} The training and evaluation of \method are carried out through conversations with a user simulator. Although this approach can provide high quality simulations for the conversation~\cite{lei2020conversational,zhang2020evaluating,zhang2022multiple}, the user simulator may not fully represent the dynamics and complexities of user behaviors in the real-world situations. This issue may be partially addressed by developing an LLM-based user simulator that fully utilizes the human-likeliness of LLMs to better simulate diverse and complex user behaviors.
\paragraph*{Template-Based Conversation}  
The template-based conversation simulation assumes that users can clearly express their preferences and choose specific options in multiple-choice questions. However, real-world conversations often involve more ambiguity and a wider range of responses than what is considered in our framework. To address this, we plan to integrate \planner with an LLM-based policy learning framework, such that LLM possesses more flexibility in handling diverse user responses, such as vague or out-of-vocabulary responses.

 \paragraph*{Cold-Start Issue} Although our main focus is on typical recommendation settings where users have historical interactions with items, and items are associated with attributes, we also acknowledge that there are cold-start settings where users do not have historical interaction with items. To adapt \method to the cold-start settings, we can disable the global information graph in \agent, and \method can still perform effective conversation planning according to performance of the variant w/o Global $\mathcal{G}$ in Section~\ref{sec:ablationstudy}. To adapt \method to settings without predefined attributes, we can perform clustering over the items’ meta data (e.g., textual descriptions, item titles) to identify the attribute types and values.

\paragraph*{Potential Risk} While we hope that CRS can provide personalized and user-friendly recommendations if correctly deployed, we also acknowledge that unintended uses of CRS may pose concerns on fairness and bias issues \cite{shen2022unintended}, which may be a potential risk for CRS but can be mitigated with debiasing algorithms as in the literature~\cite{fu2021popcorn,lin2022quantifying}. 
\section{Ethics Statement}
All datasets used in this research are from public benchmark open-access datasets, which are anonymized and do not pose privacy concerns.
\bibliography{custom}
\appendix
\setcounter{table}{0}
\setcounter{figure}{0}
\renewcommand{\thetable}{A\arabic{table}}
\renewcommand{\thefigure}{A\arabic{figure}}
\begin{table*}[ht]
    \centering
    \small
    \setlength{\tabcolsep}{2pt}{\begin{tabular}{lll}
    \toprule
    \textbf{Notation}&\textbf{Description}\\
    \midrule
    $\mathcal{U}, \mathcal{V}, \mathcal{Y}, \mathcal{P}$
    &the set of users, items, attribute types and attribute values\\
    $u, v, y, p$
    &the index of user, item, attribute type and attribute value\\
    $t,T$&the index of the current turn and the final turn of the conversation\\
    $\mathcal{G}$&the global information graph\\
    $\mathcal{G}^{+}_{t}, \mathcal{G}^{-}_{t}$& the user's positive feedback graph and negative feedback graph at the $t$-th turn\\
    $\mathcal{M}(u)$
    &the MDP environment for user $u$\\$\mathcal{S},\mathcal{A},\mathcal{T},\mathcal{R}, \gamma$
    & the state, action, transition, reward and discount factor in MDP\\
    $s_{t}, o_{t}, a_{t}, r_{t}$ & the state, action type, action and reward at the $t$-th turn\\
    $\mathbf{s}_{t}, \mathbf{a}_{t}$ & the representation of the state $s_{t}$, the embedding of the action $a_{t}$\\
    $\mathcal{A}_{s_t}$&The action space at the state $s_t$\\
    $\mathcal{A}_{s_t, o_{t}}$& The sub action space at the state $s_t$ after choosing the action type $o_t$\\$\mathcal{P}^{+}_{t},\mathcal{P}^{-}_{t},\mathcal{V}^{-}_{t}$
    & the accepted attribute values, the rejected attribute values and the rejected items at the $t$-th turn\\
    $\mathcal{P}^{c}_{t},\mathcal{V}^{c}_{t}$
    & the candidate attribute values and the candidate items at the $t$-th turn\\
    $\pi_{\phi}(o_{t}|s_t)$ & the policy network that decides the action type $o_{t}\in\{\mathtt{ask},\mathtt{rec}\}$ from the current state $s_t$\\
    $Q_{\theta}(a_t| s_t, o_t)$ & the Q-network that decides the specific action $a_t$ according to the action type $o_t$\\
    $q(s_t, o_t)$ & the expected future reward of selecting action type $o_t$ at the state $s_t$\\
    $\tau^{(u)}_{i}$ & the trajectory from the $i$-th simulation for the user $u$\\
    $R_t(\tau^{(u)}_{i})$ & the cumulative reward of trajectory $\tau^{(u)}_{i}$ from turn $t$ to the final turn $T$\\
    $V(s_t)$ & the visit count of node $s_t$ during MCTS simulations\\
    $f(s_t, o_t)$ & the child node of $s_t$ after choosing the action type $o_t$\\
    $E$& training steps\\
    $N$& the number of simulations in MCTS\\ 
    $w$& the exploration factor in UCT\\
    \bottomrule
    \end{tabular}}
    \caption{Table of notations.}
    \label{tab:notations}
\end{table*}
\newpage
\section{Table of Notations}
\label{sec:notation}
Table~\ref{tab:notations} summarizes the notations in this paper.
\section{Illustration of the State}
\label{sec:state_illustration}
\begin{figure*}
    \centering
    \includegraphics[width=\linewidth]{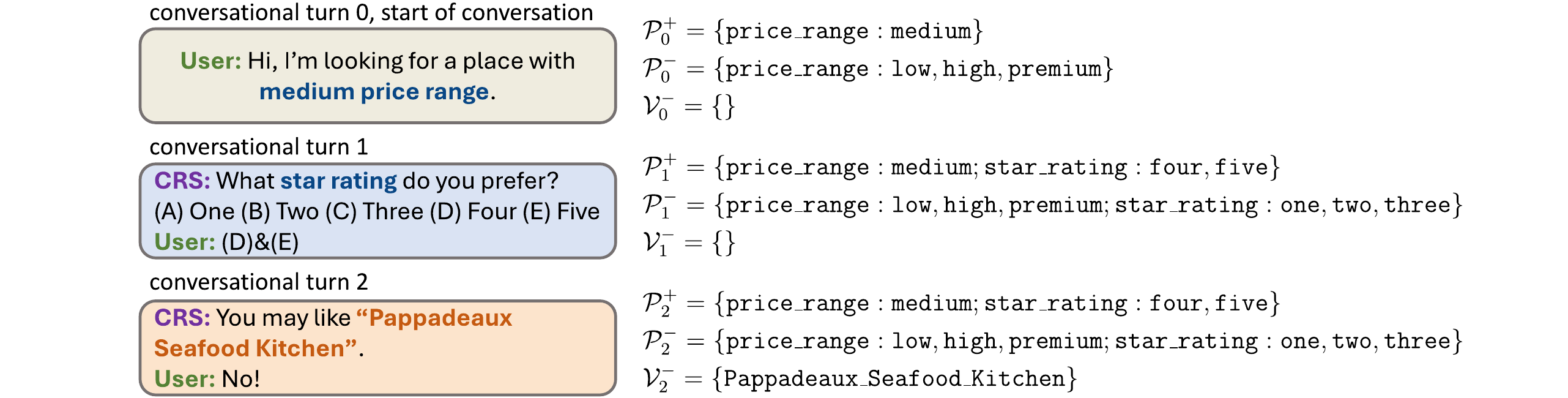}
    \caption{An illustration of the state $s_t=(\mathcal{P}^{+}_{t},\mathcal{P}^{-}_{t},\mathcal{V}^{-}_{t})$, which include all the attribute values $\mathcal{P}^{+}_{t}$ that the user has accepted, all the attribute values $\mathcal{P}^{-}_{t}$ that the user has rejected, and all the items $\mathcal{V}^{-}_{t}$ that the user has rejected until the $t$-th turn. Note that in this example, ``looking for a medium price range'' at the start of the conversation infers that all the other price ranges (low, high and premium) are not acceptable.}
    \label{fig:state_illustration}
\end{figure*}

An illustration on how to calculate the state $s_t$ is presented in Figure~\ref{fig:state_illustration}.
\section{Details of the State Encoder}
\label{technicaldetails_state_encoder}
\textbf{Global information graph encoder} captures the global relationships between similar users and items, as well as the correlations between items and attribute values from the global information graph $\mathcal{G}$. We build $\mathcal{G}$ with the following rules: an edge $e_{u,v}\in\mathcal{E}_{\mathcal{U},\mathcal{V}}$ exists between a user $u$ and an item $v$ iff. the user $u$ has interacted with item $v$, and an edge $e_{p,v}\in\mathcal{E}_{\mathcal{P},\mathcal{V}}$ exists between an attribute value $p$ and an item $v$ iff. item $v$ is associated with attribute value $v$. Next, let $\mathbf{h}^{(0)}_{u}=\mathbf{e}_{u}$, $\mathbf{h}^{(0)}_{v}=\mathbf{e}_{v}$ and $\mathbf{h}^{(0)}_{p}=\mathbf{e}_{p}$ denote the embeddings of user, item and attribute value, we adopt a multi-head Graph Attention Network (GAT) \cite{veličković2018graph,brody2022how} to iteratively refine the node embeddings with neighborhood information:
\begin{equation}
\mathbf{z}^{(l+1)}_{i}=\mathop{\Bigm|\Bigm|}\limits_{k=1}^{K}\sum_{j\in\mathcal{N}_{i}}\alpha^{k}_{ij}\mathbf{W}^{(l)}_{2,k}\mathbf{h}_{j}\nonumber
\end{equation}
\begin{gather}
\alpha^{k}_{ij}=\frac{\exp({\mathbf{a}_{k}^{(l)}}^{\top}\sigma(\mathbf{W}_{1,k}^{(l)}\mathbf{h}_{i}+\mathbf{W}_{2,k}^{(l)}\mathbf{h}_{j}))}{\sum\limits_{j^{'}\in\mathcal{N}_{i}}\exp({\mathbf{a}_{k}^{(l)}}^{\top}\sigma(\mathbf{W}_{1,k}^{(l)}\mathbf{h}_{i}+\mathbf{W}_{2,k}^{(l)}\mathbf{h}_{j^{'}}))}\nonumber\\
\mathcal{N}_{i}= \left \{
\begin{array}{ll}
    \{v\thinspace|\thinspace e_{i,v}\in\mathcal{E}_{\mathcal{U},\mathcal{V}}\},                    & \mathrm{if}\quad i\in \mathcal{U}\\
    \{v\thinspace|\thinspace e_{i,v}\in\mathcal{E}_{\mathcal{P},\mathcal{V}}\},     &\mathrm{if}\quad i\in \mathcal{P}\\
    \{u\thinspace|\thinspace e_{u,i}\in\mathcal{E}_{\mathcal{U},\mathcal{V}}\},                                 & \mathrm{if}\quad i\in\mathcal{V},\mathcal{N}_{i}\subset\mathcal{U}\\
     \{p\thinspace|\thinspace e_{p,i}\in\mathcal{E}_{\mathcal{P},\mathcal{V}}\},                                 & \mathrm{if}\quad i\in\mathcal{V},\mathcal{N}_{i}\subset\mathcal{P}\\
\end{array}
\right.,
\end{gather}
where $\mathbf{a}^{(l)}_{k}\in\mathbb{R}^{d/K}$, $\mathbf{W}^{(l)}_{1,k}\in\mathbb{R}^{({d/K})\times{d}}$,$\mathbf{W}^{(l)}_{2,k}\in\mathbb{R}^{{(d/K})\times{d}}$ are the trainable parameters for the $l$-th layer, $K$ denotes the number of attention heads, $\sigma$ denotes the LeakyReLU activation function, $\mathop{||}$ denotes the concatenation operation. For the user and attribute value node, its hidden representation of the $l{+}1$-th layer is obtain from $\mathbf{h}_{u}^{(l+1)}=\sigma(\mathbf{z}^{(l+1)}_{u})$, $\mathbf{h}_{p}^{(l+1)}=\sigma(\mathbf{z}^{(l+1)}_{p})$. While for the item node, its hidden representation of the $l{+}1$-th layer is obtained by aggregating the information from both its neighbourhood users and attribute values: $\mathbf{h}_{v}^{(l+1)}=\sigma((\mathbf{z}^{(l+1)}_{v,\mathcal{N}_{v}\subset\mathcal{U}}+\mathbf{z}^{(l+1)}_{v,\mathcal{N}_{v}\subset\mathcal{P}})/2)$. We stack $L_{g}$ layers of GATs and fetch the hidden representations $\mathbf{h}^{(L_{g})}_{u}$, $\mathbf{h}^{(L_{g})}_{v}$, $\mathbf{h}^{(L_{g})}_{p}$ at the last layer as the output of the global information graph encoder.

\textbf{Positive feedback graph encoder} captures the user's positive feedback on attribute values and their relations with candidate attribute values/items in the conversation history. For each user $u$, at the $t$-th conversational turn, we construct a local positive graph $\mathcal{G}^{+}_{t}=<(\{u\}\cup\mathcal{P}^{+}_{t}\cup\mathcal{P}^{c}_{t}\cup\mathcal{V}^{c}_{t}),\mathcal{E}^{+}_{t}>$, where the weight of the edge $\mathcal{E}^{+}_{t}(i,j)$ between node $i$ and $j$ is constructed from the following rules:
\begin{equation}
\label{edge}
    \mathcal{E}^{+}_{t}(i,j)= \left\{
    \begin{array}{ll}
    w^{(t)}_{v}, & \mathrm{if} \quad i\in\mathcal{U}, j\in\mathcal{V}
       \\
       1,  & \mathrm{if}\quad i\in\mathcal{V}, j\in\mathcal{P}\\
       1, & \mathrm{if}\quad i\in\mathcal{U}, j\in\mathcal{P}^{+}_{t}\\
       0, & \mathrm{otherwise}
    \end{array}
\right.,
\end{equation}
where $w^{(t)}_{v}=\mathrm{sigmoid}(\mathbf{e}^{\top}_{u}\mathbf{e}_{v}+\sum_{p\in\mathcal{P}^{+}_{t}}\mathbf{e}^{\top}_{v}\mathbf{e}_{p}-\sum_{p\in\mathcal{P}^{-}_{t}}\mathbf{e}^{\top}_{v}\mathbf{e}_{p})$ denotes the dynamic matching score of the item $v$ at the current conversational turn $t$. Next, let $\mathbf{e}^{(0)}_{u}=\mathbf{e}_{u}$, $\mathbf{e}^{(0)}_{v}=\mathbf{e}_{v}$ and $\mathbf{e}^{(0)}_{p}=\mathbf{e}_{p}$ denote the embeddings of user, item and attribute value, we then adopt a Graph Convolutional Network (GCN) \cite{kipf2016semi} to propagate the message on the current dynamic graph, and calculate the hidden representation of the node at the $l{+}1$-th layer as follows:
\begin{equation}
\label{graphconv}
\mathbf{e}^{\!(l\!+\!1\!)}_{i}{=}\sigma(\!\sum_{\!\{j|\mathcal{E}^{+}_{t}(i,j)>0\}\!}\!\frac{\mathbf{W}_{a}^{(l)}\mathbf{e}^{(l)}_{j}}{\sqrt{\sum\limits_{\hat{j}}\!\mathcal{E}^{+}_{t}\!(i,\hat{j})\!\sum\limits_{\hat{j}}\!\mathcal{E}^{+}_{t}\!(j,\hat{j})}}+\mathbf{e}^{(l)}_{i}),
\end{equation}
\noindent where $\mathbf{W}_{a}^{(l)}\in\mathbb{R}^{d{\times}d}$ are the trainable parameters for the $l$-th layer, $\sigma$ denotes the LeakyReLU activation function. We stack $L_{a}$ layers of GCNs and fetch the hidden representation $\mathbf{e}^{(L_{a})}_{u}$, $\mathbf{e}^{(L_{a})}_{v}$,$\mathbf{e}^{(L_{a})}_{p}$ at the last layer as the output of the positive feedback graph encoder.

\textbf{Negative feedback graph encoder} captures the user's negative feedback on attribute values and their negative correlations with candidate attribute values/items in the conversation history. Similar to Eq.~\ref{edge}, for each user $u$, we construct a local negative graph $\mathcal{G}^{-}_{t}=<\{u\}\cup\mathcal{P}^{-}_{t}\cup\mathcal{V}^{-}_{t}\cup\mathcal{P}^{c}_{t}\cup\mathcal{V}^{c}_{t},\mathcal{E}^{-}_{t}>$, where the weight of the edge $\mathcal{E}^{(t)}_{i,j}$ between node $i$ and node $j$ is constructed from the following rules:
\begin{equation}
    \mathcal{E}^{-}_{t}(i,j)= \left\{
    \begin{array}{ll}
        w^{(t)}_{v}, & \mathrm{if} \quad i\in\mathcal{U}, j\in\mathcal{V}\\
        1,  & \mathrm{if}\quad i\in\mathcal{V}, j\in\mathcal{P}\\
        1, & \mathrm{if}\quad i\in\mathcal{U}, j\in\mathcal{P}^{-}_{t}\\
        0, & \mathrm{otherwise}
    \end{array}
\right..
\end{equation}
We then stack $L_{n}$ layers of GCNs similar to Eq.~\ref{graphconv}, and fetch the hidden representation $\mathbf{e}^{(L_{n})}_{u}$, $\mathbf{e}^{(L_{n})}_{v}$,$\mathbf{e}^{(L_{n})}_{p}$ at the last layer as the output of the negative feedback graph encoder.

\textbf{Transformer-based aggregator} fuses the information from the graph encoders, and captures the sequential relationships among items and attribute values mentioned in the conversation history. Specifically, for the accepted/rejected attribute values/items at previous conversational turns, we first project the accepted ones and rejected ones into different spaces to distinguish between the positive and the negative feedbacks:
\begin{gather}
\mathbf{e}^{'}_{p}=\mathbf{W}_{a}\mathbf{e}^{(L_{a})}_{p}{+}\mathbf{b}_{a}\thinspace\thinspace\mathrm{or}\thinspace\thinspace\mathbf{e}^{'}_{p}=\mathbf{W}_{n}\mathbf{e}^{(L_n)}_{p}{+}\mathbf{b}_{n}\nonumber\\
\mathbf{e}^{'}_{v}=\mathbf{W}_{n}\mathbf{e}^{(L_n)}_{v}{+}\mathbf{b}_{n},
\end{gather}
where $\mathbf{W}_{a}, \mathbf{W}_{n}\in\mathbb{R}^{d\times d}$ and  $\mathbf{b}_{a},\mathbf{b}_{n}\in\mathbb{R}^{d}$ are trainable parameters. Next, the positive/negative feedbacks are fused with the representations from the global graph encoder with a gating mechanisms to capture the information from both the global relationships and the local conversation feedbacks:
\begin{gather}
\mathbf{v}_{p}=\mathrm{gate}(\mathbf{h}^{(L_{g})}_{p},\mathbf{e}^{'}_{p}),\thinspace\mathbf{v}_{v}=\mathrm{gate}(\mathbf{h}^{(L_{g})}_{v},\mathbf{e}^{'}_{v})\nonumber\\
\mathrm{gate}(\mathbf{x},\mathbf{y})=\xi\cdot\mathbf{x}+(1-\xi)\cdot\mathbf{y}\nonumber\\
\xi=\mathrm{sigmoid}(\mathbf{W}^{ga}_{1}\mathbf{x}+\mathbf{W}^{ga}_{2}\mathbf{y}+\mathbf{b}^{ga}),
\end{gather}
where $\mathbf{W}^{ga}_{1},\mathbf{W}^{ga}_{2}\in\mathbb{R}^{d\times d}$, $\mathbf{b}^{ga}\in\mathbb{R}^{d}$ are trainable parameters. Finally, we adopt a Transformer encoder \cite{vaswani2017attention} to capture the sequential relationships about the conversation history and obtain the current state representation $\mathbf{s}_{t}$ as follows:
\begin{equation}
\mathbf{s}_{t}=\mathrm{Meanpooling}(\mathrm{Transformer}(\mathbf{V})),
\end{equation}
where $\mathbf{V}$ is built using all the previously mentioned attribute values and items and in the conversation history: $\mathbf{V}=\{\mathbf{v}_{p}|p\in\mathcal{P}^{+}_{t}\cup\mathcal{P}^{-}_{t}\}\cup\{\mathbf{v}_{v}|v\in\mathcal{V}^{-}_{t}\}$.
\section{Transition Function}\label{transitiondetails}
Transition occurs from the current state $s_t$ to the next state $s_{t+1}$ when the user responds 
to the action $a_t$ (accepts or rejects items/attribute values). The candidate items and attribute values are updated according to the user's response.
Specifically, when the action is to ask a question on attribute values, we denote $\widehat{\mathcal{P}}^{+}_{t}$ and $\widehat{\mathcal{P}}^{-}_{t}$ as the attribute values that the user accepts or rejects at the current turn $t$, the candidate attribute value set $\mathcal{P}^{c}_{t+1}$ at the next turn $t{+}1$ is updated as $\mathcal{P}^{c}_{t+1}=\mathcal{P}^{c}_{t}\setminus(\widehat{\mathcal{P}}^{+}_{t}\cup\widehat{\mathcal{P}}^{-}_{t})$, the set of all the attribute values that the user has rejected until the $t{+}1$-th turn is updated as $\mathcal{P}^{-}_{t+1}=\mathcal{P}^{-}_{t}\cup\widehat{\mathcal{P}}^{-}_{t}$, and the set of all the attribute values that the user has accepted until the $t{+}1$-th turn is updated as $\mathcal{P}^{+}_{t+1}=\mathcal{P}^{+}_{t}\cup\widehat{\mathcal{P}}^{+}_{t}$.
When the action is to recommend items, if the user rejects all the recommended items, we denote $\widehat{\mathcal{V}}^{-}_{t}$ as the set of the recommended items at the current turn $t$ that are all rejected, and the set of all items that the user has accepted until the $t{+}1$-th turn is updated as $\mathcal{V}^{-}_{t+1}=\mathcal{V}^{-}_{t}\cup\widehat{\mathcal{V}}^{-}_{t}$; otherwise the conversation successfully finishes since the user has accepted at least one recommended item, and no state information is updated.
Finally, we update the candidate item set $\mathcal{V}^{c}_{t+1}$ at the next turn $t{+}1$ to include only those items that are still not rejected and whose attribute values have an intersection with the set of the accepted attribute values: $\mathcal{V}^{c}_{t+1}=\{v|(v{\in}\mathcal{V}_{p_{0}}{\setminus}\mathcal{V}^{-}_{t+1})\wedge(\mathcal{P}(v)\cap\mathcal{P}^{+}_{t+1}{\neq}\emptyset)\wedge(\mathcal{P}(v)\cap\mathcal{P}^{-}_{t+1}{=}\emptyset)\}$, where $\mathcal{V}_{p_{0}}$ denotes the set of items that are associated with the attribute value $p_{0}$ specified by the user at the start of the conversation.
\section{Reward Function}\label{detailreward}
Following the literature~\cite{lei2020conversational,zhang2022multiple}, for different conversation scenarios, we consider five kinds of immediate rewards at given conversational turn: (1) $r^{+}_{\rm{rec}}=1$: a large positive value when the user accepts a recommended item; (2) $r^{+}_{\rm{ask}}=0.01$: a small positive value when the user accepts an attribute value asked by \agent; (3) $r^{-}_{\rm{rec}}=-0.1$: a negative value when the user rejects a recommended item; (4) $r^{-}_{\rm{ask}}=-0.1$: a negative value when the user rejects an attribute value asked by \agent and (5) $r_{\rm{quit}}=-0.3$: a large negative value if the conversation reaches the maximum number of turns $T_{\text{max}}$. In addition, since we follow the multi-choice MCR setting, we sum up the positive and negative rewards for multiple attribute values specified in a multiple-choice question: $r_{t}=\sum_{\widehat{\mathcal{P}}^{+}_{t}}r^{+}_{\mathrm{ask}}+\sum_{\widehat{\mathcal{P}}^{-}_{t}}r^{-}_{\mathrm{ask}}.$
\section{Experimental Details}
\subsection{Dataset Details and Statistics }\label{datasetstatistics}
\begin{table}[ht]
    \centering
    \small
    \renewcommand\arraystretch{1.0} 
    \setlength{\tabcolsep}{1pt}{\begin{tabular}{lcccc}
    \toprule
    \textbf{Dataset}&\textbf{Yelp}&\textbf{LastFM}&\textbf{Amazon-}&\textbf{MovieLens}\\
    &&&\textbf{Book}&\\
    \midrule
    \#Users&27,675&1,801&30,291&20,892\\
    \#Items&70,311&7,432&17,739&16,482\\
    \#Interactions&1,368,609&76,693&478,099&454,011\\
    \#Attribute Values &590&8,438&988&1,498\\
    \#Attribute types&29&34&40&21\\
    \midrule
    \#Entities&98,576&17,671&49,018&38,872\\
    \#Relations&3&4&2&2\\
    \#Triplets&2,533,827&228,217&565,069&380,016\\
    \bottomrule
    \end{tabular}}
    \caption{Statistics of datasets after preprocessing.}
    \label{dataset_statistics}
\end{table}
We evaluate \method on four public benchmark recommendation datasets: Yelp \cite{lei2020estimation}, LastFM \cite{lei2020estimation}, Amazon-Book \cite{mcauley2015image,he2016ups} and MovieLens \cite{harper2015movielens}. The statistics of the datasets after preprocessing are presented in Table~\ref{dataset_statistics} and the details of the datasets are introduced as follows:
\begin{itemize}[noitemsep,nolistsep,leftmargin=*]
    \item \textbf{Yelp}\footnote{\url{https://www.yelp.com/dataset/}}: This dataset contains users' reviews on business venues such as restaurants and bars. \citet{lei2020estimation} build a 2-layer taxonomy for the original attribute values for this dataset, and we adopt the categories from the first layer as attribute types, the categories from the second layer as attribute values.
    \item \textbf{LastFM}\footnote{\url{https://grouplens.org/datasets/hetrec-2011/}}: This dataset contains users' listen records for music artists from an online music platform. Following the literature~\cite{lei2020conversational,zhang2022multiple}, we adopt a clustering algorithm to categorize the original attribute values into 34 attribute types.
    \item \textbf{Amazon-Book}\footnote{\url{https://jmcauley.ucsd.edu/data/amazon/}}: The Amazon review dataset \cite{mcauley2015image,he2016ups} is a large-scale collection of online shopping data featuring users' product reviews across various domains. We select the book domain from this collection. Following the literature~\cite{wang2019kgat} we choose relations and entities within the knowledge graph as attribute types and attribute values, and only retain entities associated with at least 10 items to ensure dataset quality.
    \item \textbf{MovieLens}\footnote{\url{https://grouplens.org/datasets/movielens/}}\cite{harper2015movielens}: This dataset contains users' activities in an online movie recommendation platform. We use the version with about 20M interactions, select entities and relations within the knowledge graph as attribute values, and only retain the user-item interactions with the user's ratings greater than 3 to ensure the quality of the dataset.
\end{itemize}
\subsection{Details of the User Simulator}\label{detailusersimulator}
Training and evaluating CRS with real user interactions can be impractically expensive at scale. To address this issue, we follow the literature~\cite{lei2020estimation,deng2021unified,zhang2022multiple,zhao2023multi,qian2023hutcrs} and simulate a conversation session for each observed user-item set interaction pair $(u, \mathcal{V}(u))$ in the dataset.
In each simulated conversation, we regard an item $v_{i}\in\mathcal{V}(u)$ as the ground-truth target item.
Each conversation is initialized with a user specifying preference on an attribute value $p_{0}$ that this user clearly prefers, which is randomly chosen from the shared attribute values of all items in $\mathcal{V}(u)$.
As the conversation continues, in each turn, the simulated user feedback follows these rules:
(1) when the CRS asks a question, the user will only accept attribute values associated with any item in $\mathcal{V}(u)$ and reject others;
(2) when the CRS recommends a list of items, the user will accept it only if at least one item in $\mathcal{V}(u)$ is in the recommendation list;
(3) the user will become impatient after $T_{\text{max}}=15$ turns and quit the conversation.

\subsection{Details of hDCG Calculation}\label{hdcgdetail}
Normalized Discounted Cumulative Gain (NDCG) is a common ranking metric to evaluate the relevance of items recommended by a system. \citet{deng2021unified} extend the NDCG metric to a two-level hierarchical version to evaluate the ranking order of the ground-truth item among the list of all the items recommended by the CRS at each conversational turn. A higher value implies that the ground-truth item has a higher ranking, and hence indicates a better performance for the CRS. The hierarchical normalized Discounted Cumulative Gain$@(T,K)$ (hDCG$@(T,K)$) is calculated as follows:
\begin{equation}
    \begin{aligned}
&hDCG@(T,K)=\sum^{T}_{t=1}\sum^{K}_{k=1}r(t,k)\bigg[\frac{1}{\log_{2}(t+2)}\\
    &+\left(\frac{1}{\log_{2}(t+1)}-\frac{1}{\log_{2}(t+2)}\right)\frac{1}{\log_{2}(k+1)}\bigg],
    \end{aligned}
\end{equation}
where $T$ represents the number of conversational turns, $K$ represents the number of items recommended at each turn, $r(t,k)$ denotes the relevance of the result at turn $t$ and position $k$. Since we have a maximum of $T_{\text{max}}$ conversational turns, and the CRS may recommend a maximum number of $K_{v}$ items, we report the metric $hDCG(T,K)$ where $T=T_\text{max}$ and $K=K_{v}$.
\subsection{Details of Baseline Methods}\label{baselinedetail}
\begin{figure*}
    \centering
    \includegraphics[width=\linewidth]{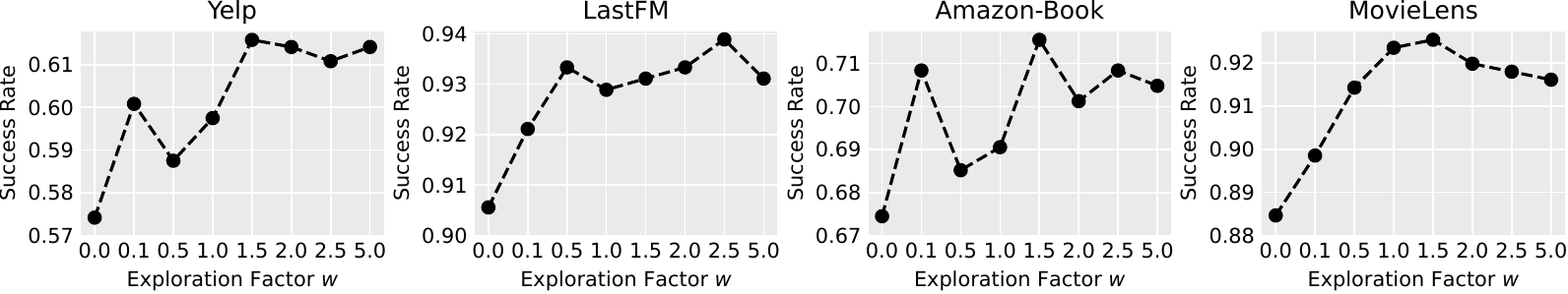}
    \caption{Success rate under different exploration factor $w$.}
    \label{fig:exploration_factor_c}
\end{figure*}
\begin{figure*}
    \centering
    \includegraphics[width=\linewidth]{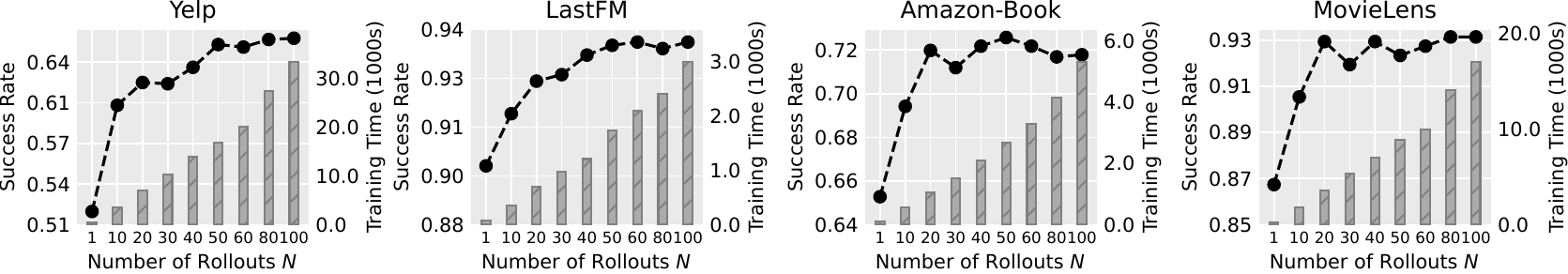}
    \caption{Success rate and training time (per 100 gradient descent steps) under different rollout number $N$. The dotted lines represent the success rate, and the bar charts represent the training time.}
    \label{fig:rollouts}
\end{figure*}
For a comprehensive evaluation, we compare \method with the following baselines:
\begin{itemize}[noitemsep,nolistsep,leftmargin=*]
    \item \textbf{Max Entropy} \cite{lei2020estimation}. This method chooses to ask for attribute values with the maximum entropy among candidate items, or chooses to recommend the top-ranked items with certain probability.
    \item \textbf{Abs Greedy} \cite{christakopoulou2016towards}. This method only recommends items in each turn without asking questions. If the recommended items are rejected, the model updates by treating them as negative samples.
    \item \textbf{CRM} \cite{sun2018conversational}. This method adopts a policy network to decide when and what to ask. As it is originally designed for single-turn CRS, we follow \citet{lei2020estimation} to adapt it to the MCR setting.
    \item \textbf{EAR} \cite{lei2020estimation}. This method designs a three-stage strategy to better converse with users. It first builds predictive models to estimate user preferences, then learns a policy network to take action, and finally updates the recommendation model with reflection mechanism.
    \item \textbf{SCPR} \cite{lei2020interactive}. This method models CRS as an interactive path reasoning problem over the knowledge graph of users, items and attribute values. It leverages the graph structure to prune irrelevant candidate attribute values and adopts a policy network to choose actions.
    \item \textbf{UNICORN} \cite{deng2021unified}. This method designs a unified CRS policy learning framework with graph-based state representation learning and deep Q-learning.
    \item \textbf{MCMIPL} \cite{zhang2022multiple}. This method develops a multi-choice questions based multi-interest policy learning framework for CRS, which enables users to answer multi-choice questions in attribute combinations.
    \item \textbf{HutCRS} \cite{qian2023hutcrs}. This method proposes a user interest tracking module integrated with the decision-making process of the CRS to better understand the preferences of the user.
    \item \textbf{CORE} \cite{jin2023lending}. This method is a Large Language Model (LLM)-powered CRS chatbot with user-friendly prompts and interactive feedback mechanisms.
\end{itemize}
\subsection{Implementation Details}\label{implementationdetailourmethod}
Following the literature~\cite{zhang2022multiple} for a more realistic multi-choice setting, if \agent decides to ask, top-$K_{p}$ attribute values with the same attribute type will be asked from the candidate attribute value set $\mathcal{P}^{c}_{t}$ to form a multi-choice question; and if \agent decides to recommend, top-$K_{v}$ items will be recommended from the candidate item set $\mathcal{V}^{c}_{t}$. Following the literature~\cite{lei2020interactive,deng2021unified}, each dataset is randomly split into train, validation and test by a 7:1.5:1.5 ratio. We set the embedding dimension $d$ as 64, batch size as 128. We adopt an Adam optimizer \cite{kingma2015adam} with a learning rate of 1e-4. We set the discount factor $\gamma$ as 0.999. The memory size of experience replay is set as 10000. For the state encoder, the number of the global information graph encoder layers $L_g$ is set as 2, and both the number of the positive feedback graph encoder layer $L_a$ and the number of the negative feedback graph encoder layer $L_n$ are set as 1, the number of the Transformer-based aggregator layers are set as 2, and we follow the literature~\cite{deng2021unified,zhang2022multiple} to adopt TransE~\cite{bordes2013translating} from OpenKE~\cite{han2018openke} to pretrain the node embeddings with the training set. Following the literature~\cite{lei2020estimation,deng2021unified}, we set the size of recommendation list $K_v$ as 10, the maximum number of turns $T_{\text{max}}=15$. We set the default exploration factor $w$ as 1.5, the default number of rollouts $N$ as 20, and variants with different $w$ and $N$ are explored in Section~\ref{sec:hyperparameter_sensitivity}.
\begin{figure*}[htbp]
    \centering
    \begin{subfigure}[t]{0.48\textwidth}
    \centering
    \includegraphics[width=\textwidth]{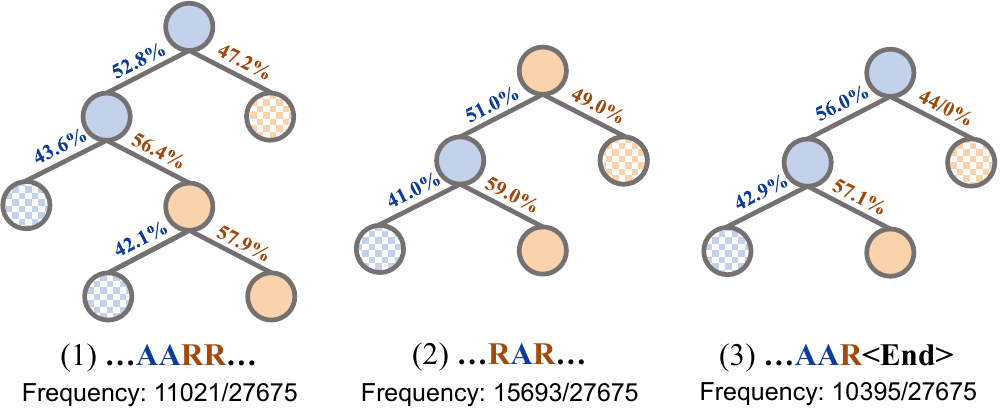}
    \caption{Yelp}
    \label{fig:action_strategy_yelp}
    \end{subfigure}\hfill
    \begin{subfigure}[t]{0.48\textwidth}
    \centering
    \includegraphics[width=\textwidth]{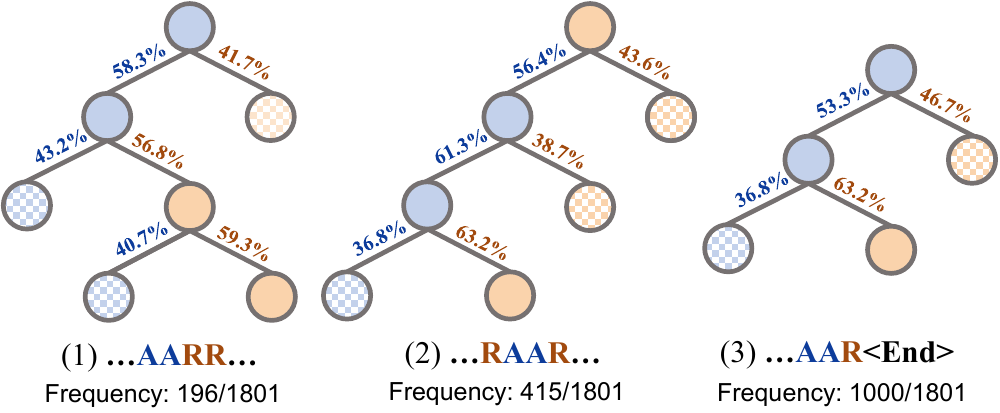}
    \caption{LastFM}
    \label{fig:action_strategy_LastFM}
    \end{subfigure}\hfill
    \begin{subfigure}[t]{0.48\textwidth}
    \centering
    \includegraphics[width=\textwidth]{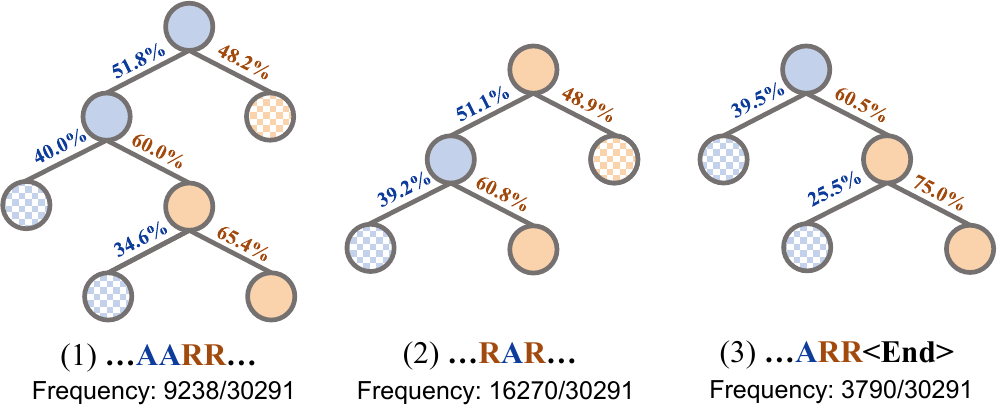}
    \caption{Amazon-Book}
    \label{fig:action_strategy_Amazon-Book}
    \end{subfigure}\hfill
    \begin{subfigure}[t]{0.48\textwidth}
    \centering
    \includegraphics[width=\textwidth]{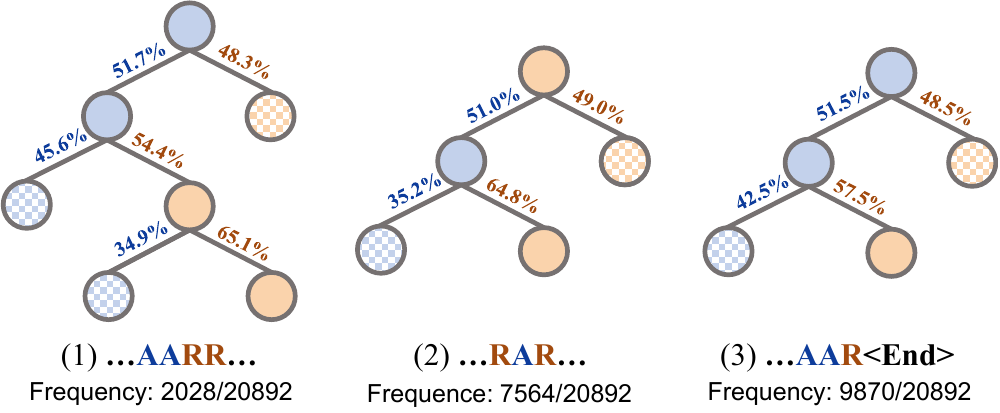}
    \caption{MovieLens}
    \label{fig:action_strategy_MovieLens}
    \end{subfigure}\hfill
    \caption{Common action strategies identified on four datasets. \textcolor[HTML]{0D3C9B}{A} stands for $\mathtt{ask}$ while \textcolor[HTML]{A24A0E}{R} stands for $\mathtt{rec}$. The probability of $\mathtt{ask}$ or $\mathtt{rec}$ at each node and the frequency (measured by \# of an action strategy$/$\# of test users in this dataset) of each action strategy are also shown in the figure. The solid circle denotes the action type that is more likely to be selected, and the shadowed circle denotes the action type that is less likely to be selected.}
    \label{fig:action_strategy}
\end{figure*}
\subsection{Hyper-Parameter Sensitivity}
\label{hyperparameter_sensitivity}
\paragraph*{Exploration and Exploitation}
The exploration factor $w$ controls the balance between exploration and exploitation. To study its impact, we set $w$ from 0.0 (exploitation only) to 5.0 (mostly favours exploration) and plot the success rate in Figure~\ref{fig:exploration_factor_c}. We find that the performance remains stable and satisfactory with high exploration, but drops with only exploitation. This is probably because our conversational search tree has a very small search space ($\mathtt{ask}$ and $\mathtt{rec}$), so high exploration does not incur much cost and also ensures thorough evaluation of different action strategies, while high exploitation may prevent the conversational search tree from discovering the optimal action strategy and lead to myopic conversational planning.

\paragraph*{Influence of MCTS rollouts}
To study the influence of MCTS rollouts, we set $N$ from 1 (equivalent to disabling MCTS, as there is no selection and reward back-propagation when $N=1$) from 50 and plot the success rate and training time (on a single Tesla V100 GPU) in Figure~\ref{fig:rollouts}. Unsurprisingly, we find that more rollouts increase the chance of discovering the optimal trajectory and lead to better performance, while disabling MCTS shows the worst performance. Nevertheless, we should also note that more rollouts bring additional computational cost, and setting $N=20$ can achieve a good trade-off between efficiency and performance.
\begin{figure*}[htbp]
    \centering
    \includegraphics[width=\textwidth]{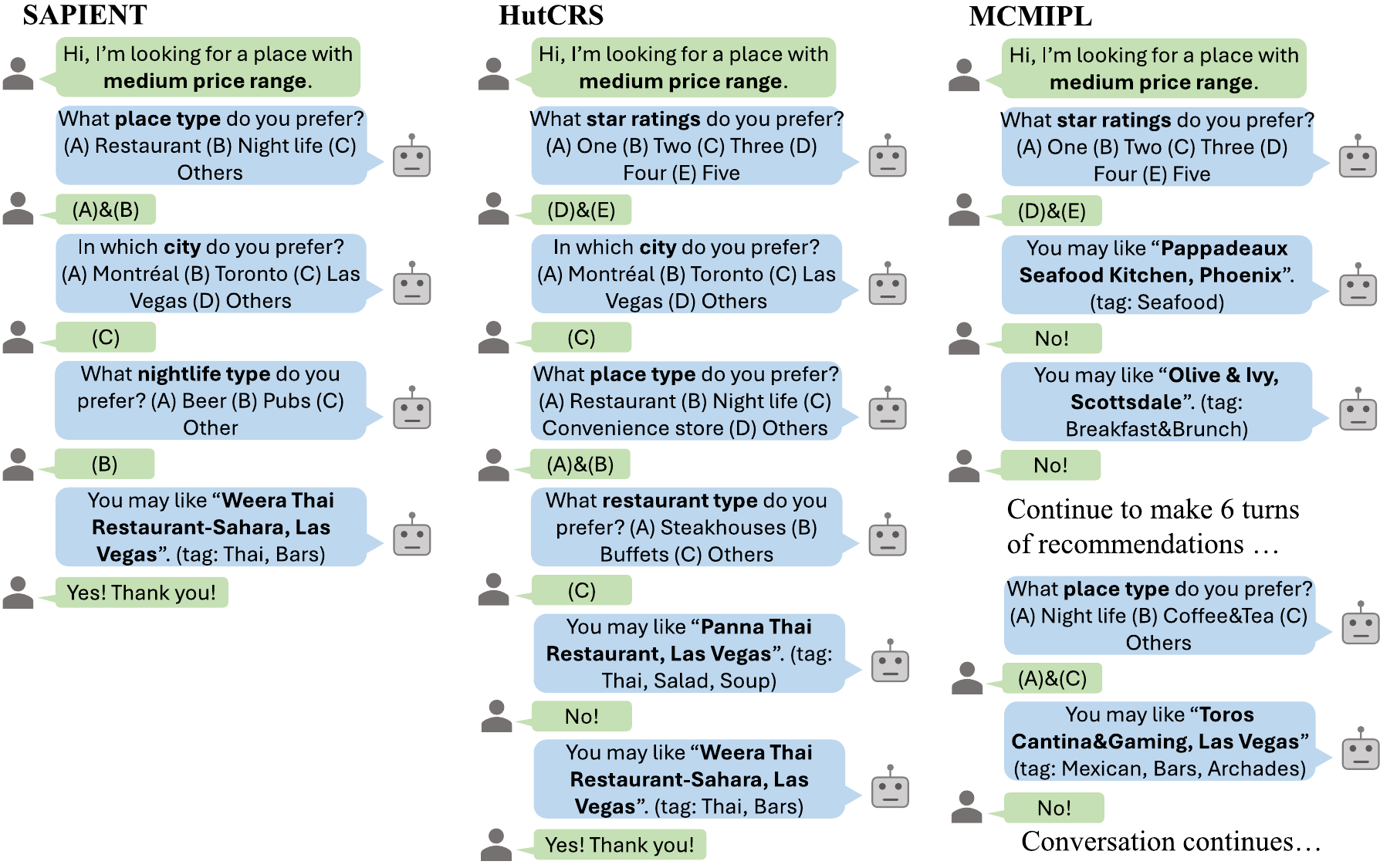}
    \caption{A case study of a user looking for a nightlife venue from the Yelp dataset.}
    \label{fig:case_study}
\end{figure*}
\subsection{Additional Analysis on Action Strategies}\label{actionstrategy}
To gain insight into the strategic planning capability of \method, we identify some typical action strategies of \method in Figure~\ref{fig:action_strategy} that are helpful for information seeking and recommendation success in the conversation. We denote \textbf{\textcolor[HTML]{0D3C9B}{A}} as the action type $\mathtt{ask}$, \textbf{\textcolor[HTML]{A24A0E}{R}} as the action type $\mathtt{rec}$, \textbf{...} as the continuation of the conversation, \textbf{<end>} as the success of the conversation, and we find the following common action strategies:
\begin{itemize}[noitemsep,nolistsep,leftmargin=*]
    \item \textbf{...\textcolor[HTML]{0D3C9B}{A}\textcolor[HTML]{0D3C9B}{A}\textcolor[HTML]{A24A0E}{R}\textcolor[HTML]{A24A0E}{R}...}: This action strategy occurs frequently during the conversation. \agent first asks the user two questions consecutively to gather crucial information on user preference, and then quickly narrows down the candidate item list by making two targeted recommendation attempts. This strategy is highly effective because it allows the \agent to tailor its recommendations to the user's preferences based on the key information obtained from the two questions. Furthermore, based on the user's feedback from the two recommendation attempts, \agent can promptly reflect upon its action strategy and make necessary adjustments to its assessment of the user's interests, thereby improving the recommendation success rate for future turns.
    \item \textbf{...\textcolor[HTML]{A24A0E}{R}\textcolor[HTML]{0D3C9B}{A}\textcolor[HTML]{0D3C9B}{A}\textcolor[HTML]{A24A0E}{R}...} and \textbf{...\textcolor[HTML]{A24A0E}{R}\textcolor[HTML]{0D3C9B}{A}\textcolor[HTML]{A24A0E}{R}...}: These are also two frequent action strategies during the conversation. In cases where an initial recommendation attempt fails, \agent will adeptly adjust the action strategy by asking one or two additional questions to better understand the user's preference, ensuring that subsequent recommendations are more aligned with the user's needs. Interestingly, we find that on the LastFM dataset, \agent tends to ask two additional questions, while on the other datasets, \agent typically asks only one additional question. This is probably because the LastFM dataset has a very large number of attribute values, so two additional questions are required to fully clarify the user's preference.
    \item \textbf{...\textcolor[HTML]{0D3C9B}{A}\textcolor[HTML]{A24A0E}{R}\textcolor[HTML]{A24A0E}{R}<end>} and \textbf{...\textcolor[HTML]{0D3C98}{A}\textcolor[HTML]{0D3C98}{A}\textcolor[HTML]{A24A0E}{R}<end>}: These two strategies occur frequently at the end of the conversation. Once \agent has gathered sufficient information about the user's preferences, it is able to reach successful recommendation with only one or two attempts. This strategy enables \agent to swiftly hit the target item, thereby shortening the conversation and reducing repeated recommendations.
\end{itemize}
\subsection{Case Study}
\label{casestudy}
We provide a case study of a randomly sampled user from the Yelp dataset in Figure~\ref{fig:case_study} to demonstrate how \method achieves strategic conversational planning. 
The user, who has previously visited some Thai restaurants, is now looking for a nightlife venue in this conversation. 
\method quickly grasps user preference by asking only three questions and makes a successful recommendation on the first attempt. By comparison, HutCRS can also make successful recommendations but requires more questions and recommendation attempts, while MCMIPL repeatedly makes failed recommendations.
Owing to the global information graph encoder, \agent can infer user preferences from historical visits (e.g., the user's preference on Thai food) without the need for explicitly queries, thus reducing conversational turns and improving the comprehension of user preferences. 
Moreover, the progression from broad questions (e.g., place type) to specific questions (e.g., nightlife type) exemplifies how \method strategically plans conversations and asks information-seeking questions, with the policy network focusing on conversation strategy planning and the Q-network specializing in the precise assessment of the attribute values and the items.
This design helps \agent to quickly narrow down candidate items and improve the recommendation success rate.
\subsection{Additional Analysis on Efficiency, Model Size, and Budget}
Although training \method requires conducting multiple simulated rollouts for each user, we find that such design will not significantly compromise efficiency compared with the baseline CRS methods. 
Under the same training pipeline with a single Tesla V100 GPU, \method with 20 rollouts per user takes 698 seconds per 100 gradient descent steps on the LastFM dataset and 1049 seconds per 100 gradient descent steps on the Amazon-Book dataset on average, which is about twice as slow as the two competitive baselines (HutCRS: 305 seconds/100 steps on LastFM, 465 seconds/100 steps on Amazon-Book; MCMIPL: 429 seconds/100 steps on LastFM, 548 seconds/100 steps on Amazon-Book). 
This is because the simulation process only requires forward calculation without the need for gradient backward update, so even conducting 20 rollouts per user will only reduce the training speed by half.
Moreover, \methodeff takes 397 seconds per 100 gradient descent steps on the LastFM dataset and 593 seconds per 100 gradient descent steps on the Amazon-Book dataset on average, which is highly comparable to baselines.
Furthermore, we note that during inference, the efficiency of \method is comparable with baseline methods, because no tree search is required during inference, and the number of parameters for \agent (1.30M on Yelp, 3.30M on LastFM, 2.65M on Amazon-Book, 6.48M on MovieLens) is also budget-friendly.
For these reasons, we think that it is worthwhile to introduce conversational tree search for CRS, because such design only slightly compromises efficiency during training, and the training efficiency can also be improved by adopting parallel acceleration methods for MCTS \cite{chaslot2008parallel}.
\end{document}